\documentclass[10pt,twocolumn,letterpaper]{article}

\usepackage[pagenumbers]{cvpr}

\usepackage[accsupp]{axessibility}

\PassOptionsToPackage{dvipsnames}{xcolor}
\usepackage{multirow}
\usepackage{anyfontsize}
\usepackage{xspace}
\usepackage{subcaption}
\usepackage[export]{adjustbox}
\usepackage{tabularx}

\usepackage[table]{xcolor}
\usepackage{graphicx}
\usepackage{array}
\usepackage{pifont}
\usepackage{enumitem}
\usepackage{caption}
\usepackage{tikz}
\usepackage{fontawesome5}

\newcommand{\citeMain}[1]{\cite{#1}}

\newcolumntype{Y}{>{\centering\arraybackslash}X} 
\newcommand{\rotangle}{0}

\newcommand{\myparagraph}[2][0.5]{\vspace{#1em}\noindent{\bf #2}}

\newcommand\bcos{B-cos\xspace}

\newcommand{\ours}{ALOE\xspace}

\definecolor{hirow}{HTML}{F3E6D9}

\DeclareRobustCommand{\prevreg}{\textsuperscript{\textdagger}}
\DeclareRobustCommand{\prevregnote}{\textsuperscript{\textdagger}Earlier CVPR-version checkpoint~\cite{Maser_2026_CVPR}; see Appendix~\ref{supp:sec:prevreg}.}
\newcommand{\hflogo}{%
  \raisebox{-0.18em}{\includegraphics[height=1.05em]{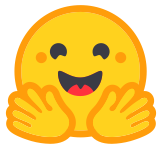}}%
}
\newcommand{\linkbadge}[3]{%
  \href{#1}{%
    \tikz[baseline=(badge.base)]\node[
      draw=black!75,
      fill=white,
      rounded corners=3pt,
      line width=0.4pt,
      inner xsep=6pt,
      inner ysep=2.25pt
    ] (badge) {\textcolor{black}{#2\hspace{0.35em}\small #3}};%
  }%
}

\newcommand{\emoji}[2][3.0ex]{%
  \raisebox{-0.2ex}{\includegraphics[height=#1]{#2}}%
}

\definecolor{mygray}{gray}{0.75}

\newcommand{\RowStyle}{}
\newcolumntype{L}{>{\RowStyle}l}
\newcolumntype{R}{>{\RowStyle}r}
\newcommand{\setrowstyle}[1]{\gdef\RowStyle{#1}}

\newcommand{\cmark}{\textcolor{green!60!black}{\ding{51}}}
\newcommand{\xmark}{\textcolor{red!70!black}{\ding{55}}}

\usepackage{booktabs}
\usepackage{graphicx}
\usepackage{wrapfig}

\definecolor{cvprblue}{rgb}{0.21,0.49,0.74}
\usepackage[pagebackref,breaklinks,colorlinks,allcolors=cvprblue]{hyperref}

\usepackage{url}

\def\paperID{}
\def\confName{CVPR}
\def\confYear{2026}

\title{Align Once to Explain~\texorpdfstring{\emoji{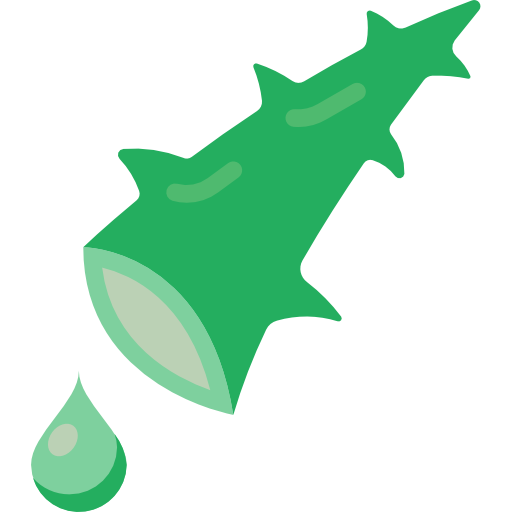}}{[aloe]}: Feature Alignment for Scalable B-cosification of Foundational Vision Transformers}

\author{Raphael Maser$^{*}$,
Siddhartha Gairola$^{*}$, Sukrut Rao, Bernt Schiele \\
Max Planck Institute for Informatics, Saarland Informatics Campus, Saarbrücken, Germany \\
{\tt\small \{raphael.maser,siddhartha.gairola,sukrut.rao,schiele\}@mpi-inf.mpg.de} \\[0.55em]
{\small
\linkbadge{https://rmaser.github.io/aloe_project/}{\faIcon{globe}}{Project Page} \quad
\linkbadge{https://github.com/rmaser/aloe}{\faIcon{github}}{Code} \quad
\linkbadge{https://huggingface.co/collections/rmaser/aloe}{\hflogo}{Models}}
}

\begin{document}

\twocolumn[{%
\renewcommand\twocolumn[1][]{#1}%
\maketitle
\begin{minipage}{0.6\linewidth}
  \includegraphics[width=\linewidth]{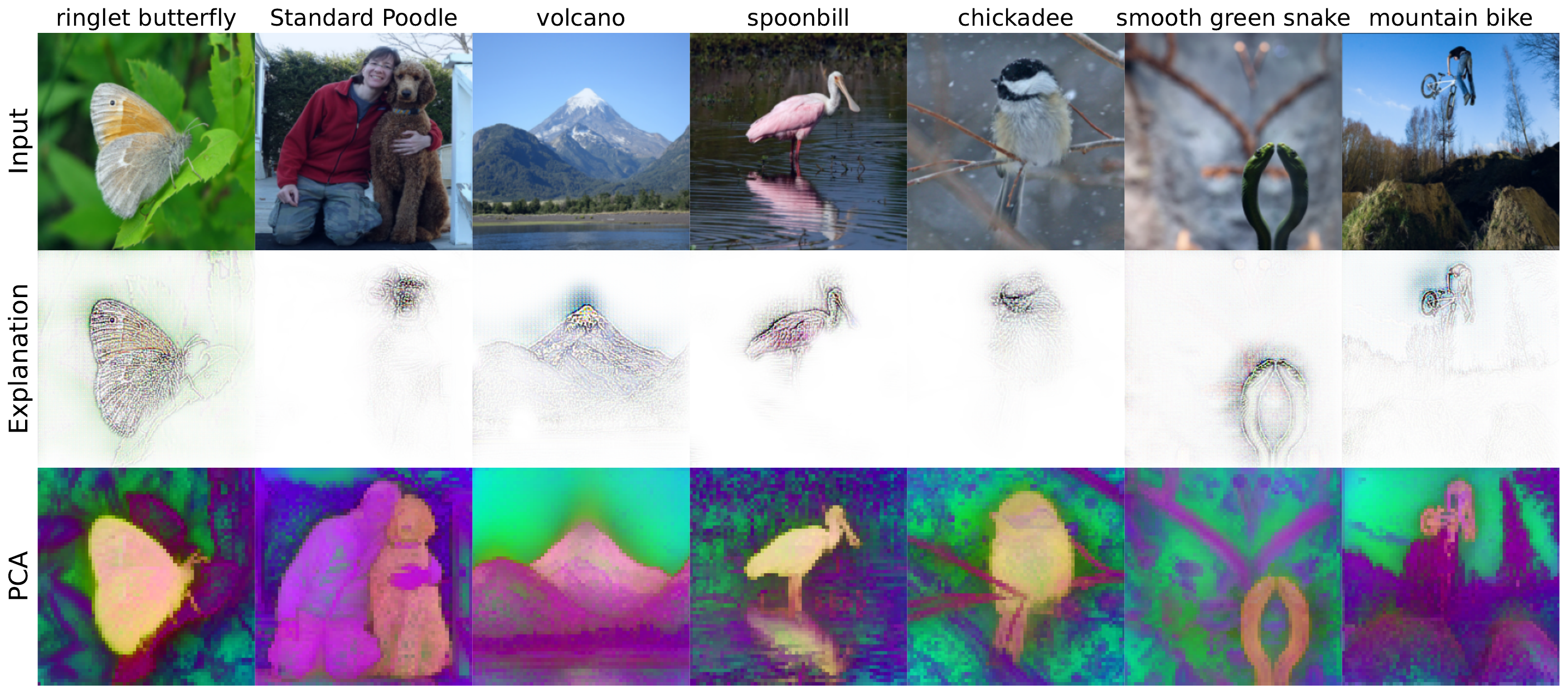}
\end{minipage}
\hfill
\begin{minipage}{0.38\linewidth}
  \includegraphics[width=\linewidth]{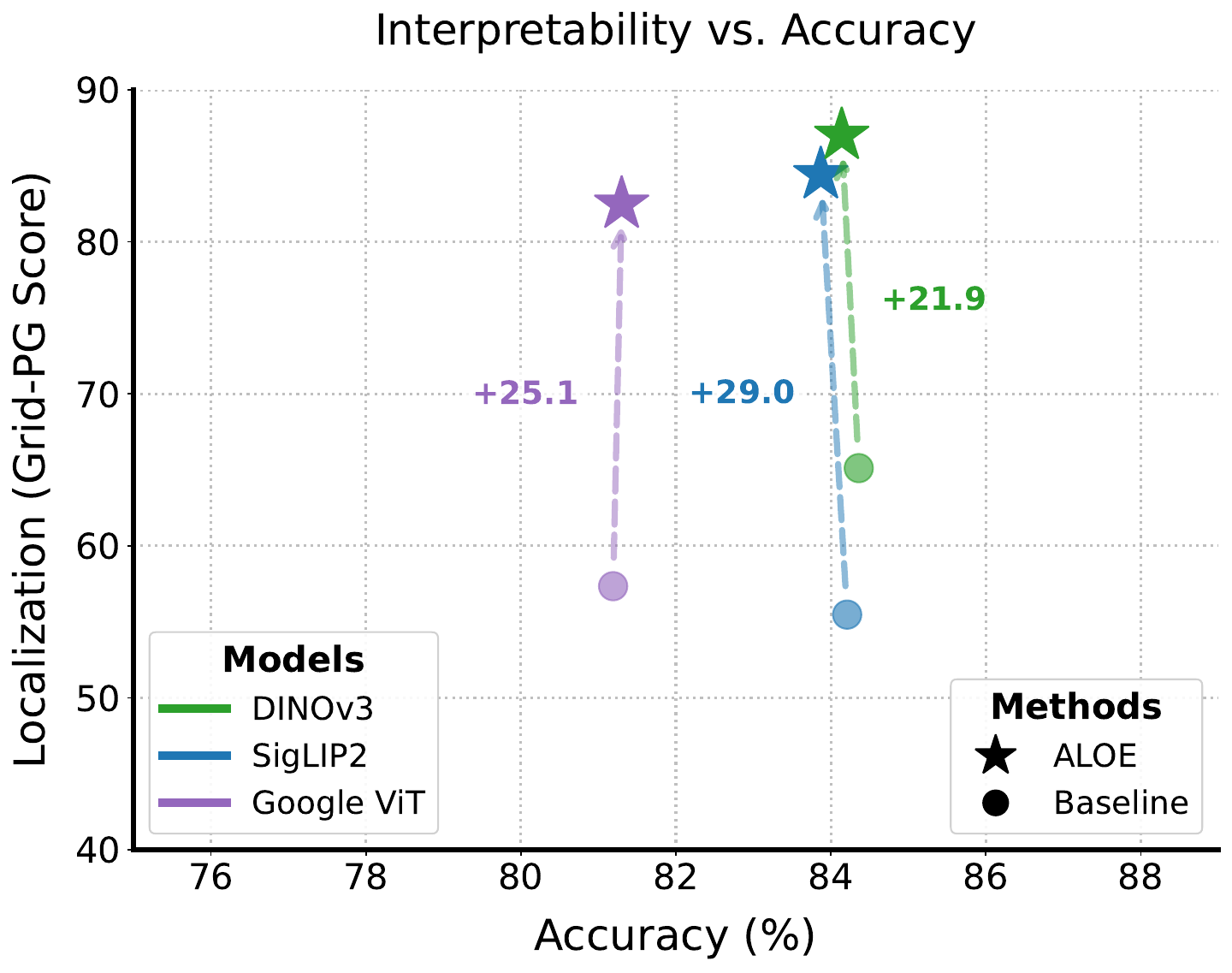}
\end{minipage}
\captionof{figure}{\textbf{\ours: performant, preserves semantics, explains faithfully.}---
\textit{Left: Explanations and representations.} \ours enables inherently interpretable B-cos: $\mathbf{W(x)}$~\citeMain{boehle2024bcos} attributions (\textbf{row 2}) that are object-centric and class-specific. The accompanying PCA visualizations (\textbf{row 3}; RGB = first three principal components of the final image representation for our \ours aligned DINOv3~\citeMain{simeoni2025dinov3} model) preserve global feature geometry—indicating aligned semantics with improved explainability (\cf~\cref{fig:pca}). \textit{Right: Performance vs.\ interpretability.} Across ViT-B/16 backbones, \ours substantially boosts localization quality on GridPG~\citeMain{boehle2021convolutional,zhang2018top}
while maintaining competitive ImageNet~\cite{imagenet} accuracy relative to the corresponding foundation models (Supervised~\citeMain{dosovitskiy2021an}, DINOv3~\citeMain{simeoni2025dinov3}, SigLIP2~\citeMain{tschannen2025siglip}). \textit{Baseline} denotes each teacher explained with AttnLRP~\citeMain{achtibat2024attnlrp}.
} 
\vspace{2em}
\label{fig:teaser}
}]

\begingroup
\renewcommand{\thefootnote}{\fnsymbol{footnote}}
\footnotetext[1]{Equal contribution.}
\endgroup
\setcounter{footnote}{0}

\begin{abstract}
Foundational vision models have become the de facto standard for many vision tasks due to their strong performance. However, they are notoriously opaque and remain hard to interpret. 
We present \textbf{\ours} (\textbf{AL}ign \textbf{O}nce to \textbf{E}xplain), a one-time, label-free feature alignment approach that efficiently converts foundational vision models into inherently interpretable \bcos~\citeMain{boehle2024bcos} variants.
Once aligned, the \bcos backbone is used as a drop-in replacement across several downstream tasks—amortizing the cost of interpretability. \ours is robust across pre-training paradigms (supervised, self-supervised, vision–language) and can preserve teacher-level generalization with orders of magnitude less data than training from scratch.
On classification, it strongly outperforms vanilla B-cosification (e.g., \textbf{+10.5 p.p.} top-1 on ImageNet for DINOv3 ViT-B/16), retains strong linear probing, $k$-NN, and zero-shot transfer competitive with foundational backbones (DINOv3~\citeMain{simeoni2025dinov3}, SigLIP2~\citeMain{tschannen2025siglip}) across diverse downstream datasets. It also preserves spatially structured features useful for dense prediction, while yielding well-localized and highly human-interpretable explanations by design.
\end{abstract}
    
\section{Introduction}
\label{sec:intro}

Large-scale vision foundation models, including self-supervised encoders in the DINO family~\citeMain{caron2021emerging, oquab2024dinov, simeoni2025dinov3} and contrastive vision–language models such as CLIP~\citeMain{radford2021learning} and SigLIP~\citeMain{zhai2023sigmoid, tschannen2025siglip}, are powerful and versatile feature extractors that achieve state-of-the-art performance across a diverse set of downstream tasks. Trained on billions of images and image–text pairs, these models serve as general-purpose building blocks across various domains, from fine-grained recognition to medical imaging, often replacing their fully supervised, domain-specific counterparts~\citeMain{simeoni2025dinov3}. However, their decision-making processes remain largely opaque, which hinders their use in sensitive and safety-critical applications. Post-hoc explanation methods aim to mitigate this issue, but can be noisy and not always faithful to the underlying model being explained~\citeMain{adebayo2018sanity,rao2022towards,adebayo2021post}.

An alternative line of work develops \emph{inherently interpretable} architectures that yield model-faithful explanations by design. They introduce architectural constraints that lead to human-interpretable, faithful summaries of the model's computations; examples include prototype-based models~\citeMain{chen2019looks, donnelly2022deformable,nauta2021neural}, dynamic linear models~\citeMain{boehle2021convolutional, boehle2022bcos}, and concept-bottleneck models~\citeMain{koh2020concept,oikarinen2023label,rao2024discover}. A prominent example is the family of B-cos Networks~\citeMain{boehle2022bcos,boehle2024bcos}, which replace linear transforms with bias-free B-cos transforms that promote weight-input alignment to enhance interpretability and enable a faithful summary of the model’s decision. To avoid training these models from scratch, recent work proposed \emph{`B-cosification'}~\citeMain{arya2024b}, an approach that efficiently transforms existing models to their B-cos variants post-hoc at a fraction of the full training cost. However, despite strong gains on fully-supervised convolutional backbones and competitive results on a CLIP ResNet~\citeMain{he2016deep} backbone, the B-cosification recipe provides only modest improvements for Vision Transformers (ViTs)~\citeMain{dosovitskiy2021an}, sometimes failing to match the performance of training from scratch. Since most modern foundation models are ViT-based, this significantly limits its practical utility and broader adoption.

\begin{figure}[t]
  \center
\centering
\begin{subfigure}{\columnwidth}
    \includegraphics[width=\columnwidth]{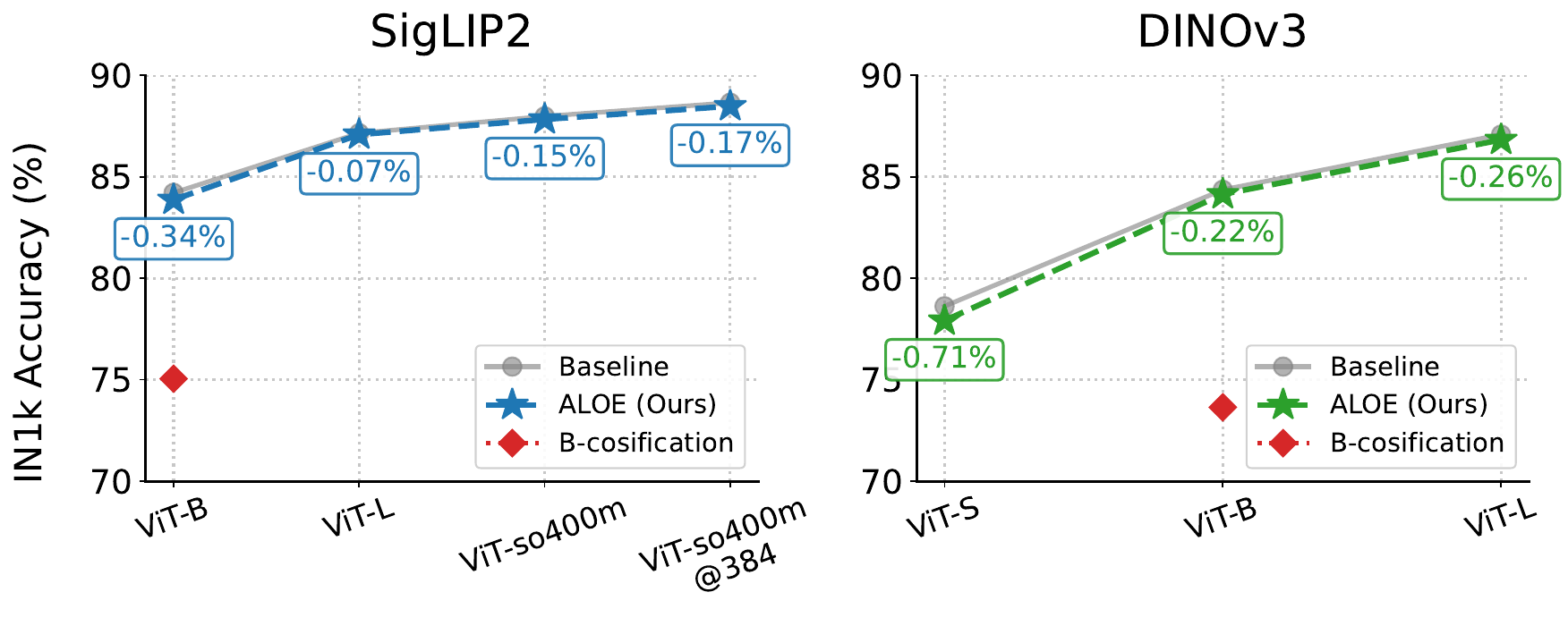}
    \caption{ImageNet-1k linear-probe accuracy (\%).}
  \end{subfigure}
  
  \vspace{0.5em}

  \begin{subfigure}{\columnwidth}
    \includegraphics[width=1\columnwidth]{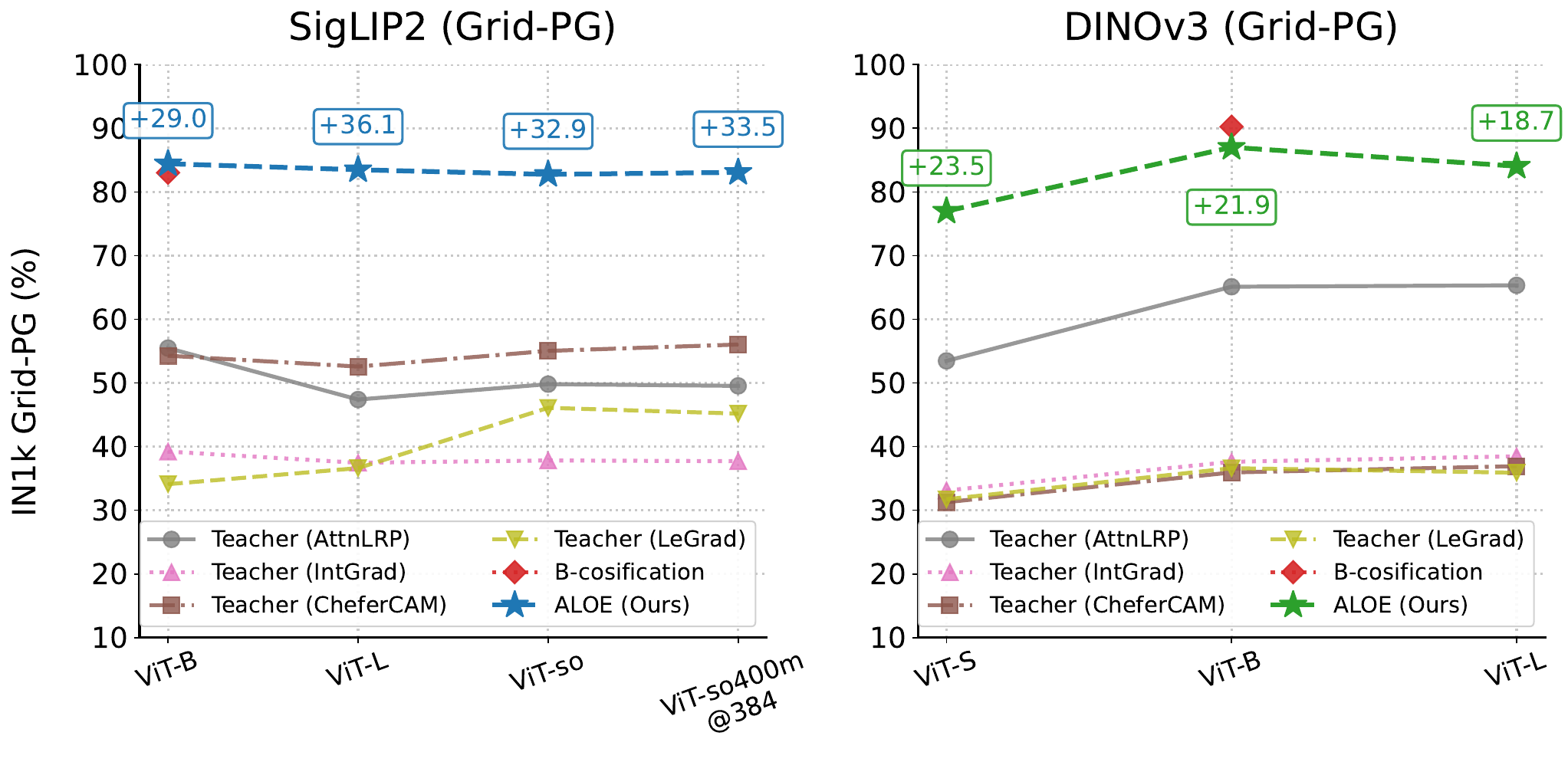}  
    \caption{Grid-PG (localization) Score (\%).}
  \end{subfigure}
  
  \caption{\textbf{ALOE vs.\ B-cosification~\citeMain{arya2024b}.} ALOE consistently outperforms vanilla B-cosification in ImageNet-1k linear-probe accuracy across model scales for both SigLIP2~\cite{tschannen2025siglip} and DINOv3~\cite{simeoni2025dinov3} (see (a)). It reaches performance close to the original SigLIP2 and DINOv3 baselines while significantly improving localization scores (Grid-PG, see (b))---even compared against different SOTA post-hoc attribution methods.}
  \label{fig:aloe_vs_bcosification}
\end{figure}

In this work, we propose a scalable approach to efficiently transform existing foundation models into \bcos variants, that is \textit{practically useful} for modern ViTs. Specifically, we explore feature-alignment methods for fine-grained matching 
between a pre-trained foundational teacher and its B-cosified student without any labeled supervision. We conduct a comprehensive study of key design choices, including which features/layers to align, alignment losses, model scales, and alignment dataset sizes. Unlike \citeMain{arya2024b}, our approach, \textbf{\ours} \textit{ (\textbf{AL}ign \textbf{O}nce to \textbf{E}xplain)} (\cref{fig:setup}), is a general-purpose recipe that requires only a single alignment of the backbone irrespective of the pre-training paradigm, such as fully-supervised, self-supervised, or contrastive vision-language training. Once aligned, the backbone can be used as a drop-in replacement across downstream tasks, amortizing the cost of interpretability.

Despite its simplicity, \ours is surprisingly effective and highly data and compute efficient. For example, while training SigLIP2~\citeMain{tschannen2025siglip} from scratch required $\approx$10B images and $\approx$12B alt-texts, \ours B-cosifies such a model while maintaining its generalization with only a small unlabeled alignment set: even 150k YFCC15M images already keep the ImageNet top-1 gap below 0.50 p.p.\ (\cref{fig:aloe_siglip_data_scaling}), and the full 15M-image alignment set reaches $83.87\%$ vs.\ $84.21\%$ for the SigLIP2 teacher (see~\cref{fig:teaser}). This makes the one-time alignment overhead clearly preferable to full retraining for achieving interpretability. Moreover, \ours consistently outperforms vanilla B-cosification~\citeMain{arya2024b} across pre-training paradigms, improving ImageNet-1k linear-probe accuracy by +8.82 to +10.50 p.p.\ for the SigLIP2 and DINOv3 ViT-B/16 models shown in \cref{fig:aloe_vs_bcosification}. Our models also deliver strong linear-probing and zero-shot performance across datasets, preserve spatially structured features useful for dense prediction, and provide faithful, well-localized explanations (\cref{fig:teaser}).

\noindent In summary, our \textbf{contributions} are:
\begin{enumerate}[label=\textbf{(\arabic*)}, leftmargin=*, itemsep=0pt, topsep=0pt]
  \item We propose \textbf{\ours} \textit{(\textbf{AL}ign \textbf{O}nce to \textbf{E}xplain)}, a universal, one-time, label-free feature-alignment approach that enables efficient transformation of foundational ViT backbones into inherently interpretable \bcos variants while preserving downstream utility.
  \item We conduct a detailed study of the design choices and provide key insights into alignment targets (layer-wise vs.\ pooled; single vs.\ multi-layer), loss families (MSE, cosine, SigLIP, InfoNCE), model scales, and alignment dataset sizes.
\end{enumerate}

\noindent We make the following \textbf{key empirical findings}:
\begin{enumerate}[label=\textbf{(F\arabic*)}, leftmargin=*, itemsep=0pt, topsep=2pt]

  \item \textbf{Generality:} \ours \bcos ViTs outperform B-cosified~\citeMain{arya2024b} baselines in classification, retain strong linear probing, \textit{k}-NN, and zero-shot transfer performance competitive with foundational backbones (DINOv3, SigLIP2), preserve spatially structured features for dense prediction, and provide faithful and well-localized explanations (e.g., \textbf{+8.82} to \textbf{+10.50 p.p.} ImageNet-1k linear-probe accuracy over~\citeMain{arya2024b} for the SigLIP2 and DINOv3 ViT-B/16 models).
  
  \item \textbf{Data efficiency:} feature alignment with as few as 150k unlabeled images keeps the SigLIP2 ImageNet top-1 gap below 0.50 p.p.\ (\cref{fig:aloe_siglip_data_scaling}); full 15M-image alignment reduces the gap to 0.34 p.p., using far less data than billion-scale pre-training.
\end{enumerate}

\noindent With \textbf{\ours} we establish a practical path to scalable trustworthy vision foundation models that are \textit{faithful by design, strong in transfer tasks, and easy to adopt.}

\section{Related work}
\label{sec:related_work}

\myparagraph{Large-scale vision foundation models}
trained on billion-scale datasets are de facto backbones for transfer learning and zero-shot tasks. These include self-supervised ViT encoders in the DINO family~\citeMain{caron2021emerging,oquab2024dinov,simeoni2025dinov3} and contrastive vision–language encoders such as CLIP~\citeMain{radford2021learning} and SigLIP/SigLIP2~\citeMain{zhai2023sigmoid,tschannen2025siglip}. Supervised ViTs pre-trained on large datasets (e.g., ImageNet21k/22k~\citeMain{kolesnikov2020big} or proprietary corpora) also serve as strong backbones for downstream usage~\citeMain{dosovitskiy2021an,kolesnikov2020big}. 
Such encoders are also used by large vision-language models and generative models together with a pre-trained LLM for multimodal understanding~\citeMain{liu2023visual,cocchi2025llava}.
In this work, we transform modern vision foundation models into an inherently interpretable \bcos backbone via a single alignment step, while preserving downstream utility.

\myparagraph{Inherently interpretable models.}
To understand deep neural networks (DNNs), post-hoc attribution methods spanning gradient-~\citeMain{bach2015pixel,shrikumar2017learning,sundararajan2017axiomatic,chefer2021generic} activation-~\citeMain{selvaraju2017grad,chattopadhyay2018grad,jiang2021layercam,wang2020score}, and perturbation-based~\citeMain{petsiuk2018rise} methods have been typically used, with the resulting explanation maps summarizing input regions contributing to the model's decision. In contrast, inherently interpretable architectures enforce structural constraints to give human-understandable, model-faithful summaries of the computation. These include prototype-based models~\citeMain{chen2019looks,donnelly2022deformable,nauta2021neural}, dynamic-linear models~\citeMain{boehle2021convolutional,boehle2022bcos}, and concept-bottleneck models~\citeMain{koh2020concept,oikarinen2023label,rao2024discover,ParchamiAraghi2025FaCTFC}. \bcos networks~\citeMain{boehle2024bcos} replace linear layers with bias-free \bcos transforms that promote weight–input alignment, producing faithful, well-localized explanations by design and have gained recent prominence~\citeMain{parchami2024good,gairola2025probe}. The cost of training such models from scratch at foundation scale motivated B-cosification~\citeMain{arya2024b}, which transforms existing networks to \bcos variants post hoc. While B-cosification showed gains on supervised CNNs and for a CLIP ResNet model, the transformation recipe falls short in performance for ViTs, limiting applicability to foundation models. We therefore propose a scalable approach to transform ViT-based foundation backbones into inherently interpretable \bcos variants.

\myparagraph{Knowledge distillation (KD)} 
transfers behavior from a pre-trained model (`teacher' $\mathcal{T}$) to a smaller model (`student' $\mathcal{S}$) using soft targets or intermediate features, often for compression.
Logit-based KD~\citeMain{hinton2015distilling} which aligns output distributions between $\mathcal{T}$ and $\mathcal{S}$ has been extended with feature-level and attention-based alignment objectives~\citeMain{zagoruyko2016paying,romero2014fitnets,park2019relational}, layer-wise and multi-stage schemes~\citeMain{zhou2024rethinking}, and token-level distillation for ViTs~\citeMain{touvron2021training}. Recent works~\citeMain{parchami2024good} also leverage explanation map similarity objectives which improve faithfulness over logit-only KD, and improve robustness under distribution shifts. To reduce reliance on annotations, label-free KD techniques use unlabeled or synthetic data~\citeMain{nayak2019zero,shin2024teacher}. 
In contrast to compression as a goal, we use a one-time, label-free \emph{feature alignment} objective---akin to KD---to convert a frozen vision foundation model into an inherently interpretable \bcos counterpart. Our universal recipe retains the teacher's general-purpose features, provides model faithful explanations, and requires orders of magnitude fewer images than full billion-scale pre-training.

\section{Background: \bcos networks}
\label{sec:background}

In this section we briefly review \bcos networks~\citeMain{boehle2024bcos} (\cref{sec:background_bcos_nets}) and the B-cosification~\citeMain{arya2024b} (\cref{sec:background_bcosification}) recipe proposed for CNNs, before presenting our approach (\cref{sec:aloe}).

\subsection{B-cos networks}
\label{sec:background_bcos_nets}

\begin{figure}[h!]
  \centering
  \includegraphics[trim=0mm 3mm 0mm 3mm, clip,width=0.8\linewidth]{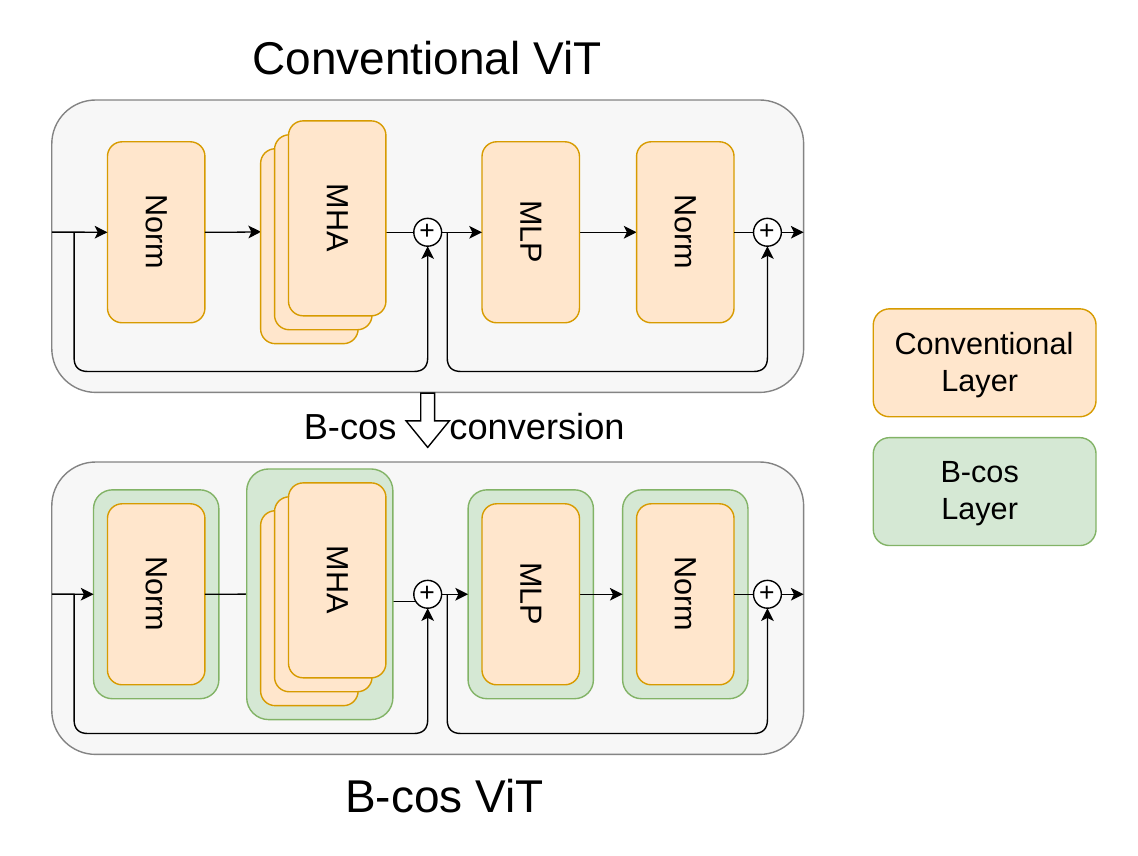}
  \caption{\textbf{Conventional vs. \bcos ViT.} We replace linear/MLP layers with \emph{bias-free} \bcos transforms and uncentered normalization, keeping self-attention and residuals intact. This yields a dynamic-linear \bcos ViT with inherently faithful explanations via $\mathbf{W}(\mathbf{x})\mathbf{x}$ (\cref{sec:background_bcos_nets}).}
  \label{fig:bcos_vit}
\end{figure}

\bcos networks~\citeMain{boehle2024bcos} are inherently interpretable architectures that provide faithful and human-interpretable explanations of model decisions. To do this, bias-free, dynamic-linear B-cos transforms are used in place of linear transforms that induce weight-input alignment across the model and provide faithful explanations that constitute an exact decomposition of the model’s computation.
The \textbf{\bcos transform}~\citeMain{boehle2022bcos} is given by:
\begin{equation}
\label{eq:bcos-transform}
{\text{\bcos}}(\mathbf{x}; \mathbf{w}) =
\left(\bigl|\cos(\mathbf{x},\mathbf{w})\bigr|^{\,B-1}\times\widehat{\mathbf{w}}\right)^{\!\top}\mathbf{x}
= \mathbf{w}(\mathbf{x})^{\top}\mathbf{x},
\end{equation}
\noindent where $B$ controls the alignment strength, $\cos$ is the cosine similarity between input $\mathbf{x}$ and weights $\mathbf{w}$, and $\widehat{\mathbf{w}}=\mathbf{w}/\lVert\mathbf{w}\rVert_2$. Stacking such units builds a \emph{dynamic-linear} network that produces an exact, faithful summary $\mathbf{y}(\mathbf{x})=\mathbf{W}(\mathbf{x})\,\mathbf{x}$, where $\mathbf{W}(\mathbf{x})$ is the effective input-dependent dynamic linear weight from the full model.
Increasing $B{>}1$ promotes weight–input alignment, making $\mathbf{W}(\mathbf{x})$ more task-relevant and interpretable. The cosine term raised to the power of $B{-}1$ provides the non-linearity needed for learning while preserving model expressivity.
B-cos variants of standard CNNs (\eg, \bcos ResNets~\citeMain{he2016deep}) can then be constructed---such models use B-cos transforms in place of convolutional and linear layers, remove activations, and remove all biases including in normalization layers.
\textbf{B-cos ViTs} (\cref{fig:bcos_vit})
 similarly replace linear layers in the patch embedding, MLP blocks, and projection heads with \bcos layers. Notably, self-attention, being already dynamic-linear~\citeMain{boehle2024bcos}, is left unchanged, as are positional embeddings.

Such models were shown to provide competitive performance while significantly improving interpretability~\citeMain{boehle2024bcos}, however, the need to train them from scratch remained a significant limitation to their adoption as compared to using already trained conventional models.

\subsection{B-cosification recipe}
\label{sec:background_bcosification} 

To alleviate the need to retrain B-cos variants of existing models, `B-cosification'~\citeMain{arya2024b} was proposed as a means to transform a pre-trained conventional model into its \bcos variant by making targeted architectural modifications followed by supervised fine-tuning for few epochs on the original task. Broadly, the transformation involves replacing linear transforms with corresponding B-cos transforms (\cref{eq:bcos-transform}) with $B=2$ and removing biases from all layers including normalization layers. In contrast to the original B-cos architecture~\citeMain{boehle2024bcos}, point-wise activation functions (ReLU/GELU) are left in and weights are not unit normalized, as it preserves the distribution of the learnt weights without harming interpretability. Following~\citeMain{boehle2024bcos}, 3-channel image inputs given by $(r,g,b)$ are preprocessed to 6-channels $(r,g,b,1{-}r,1{-}g,1{-}b)$ to allow visualizing explanations in color, with a transform to the first layer weights to maintain equivalence.

However, this recipe was designed with a focus on supervised CNNs, and was relatively ineffective for ViT-based architectures. In our work, we propose a scalable \emph{feature-alignment} approach that is highly effective for modern ViT-based foundation models.

\begin{figure*}[t]
\begin{center}
    \centering
        \includegraphics[width=0.9\linewidth]{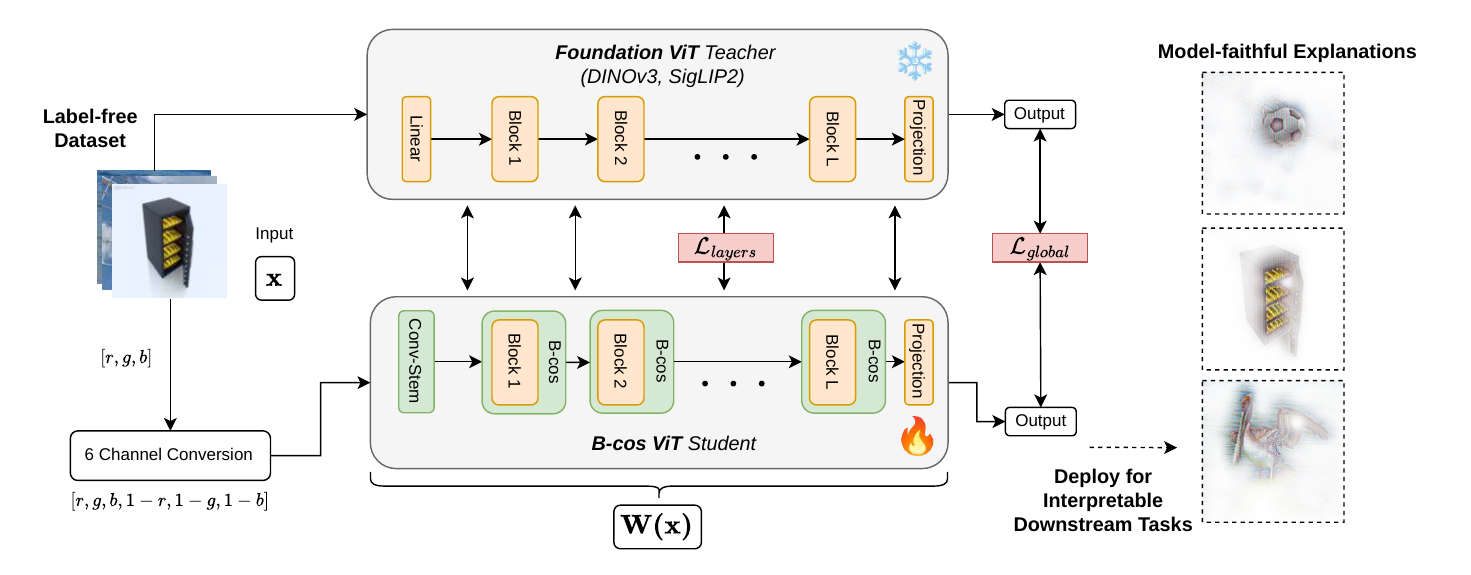}
\end{center}
    
 \caption{\textbf{Align Once to Explain.} \textit{(1) Transform to \bcos}—convert a Foundation ViT encoder into a bias-free, dynamic-linear backbone (\cref{sec:aloe_setup}).
 \textit{(2) Align once}—label-free feature alignment with cosine distance objective between foundational teacher and \bcos student, on unlabeled images (\cref{sec:aloe_alignment}). \textit{(3) Deploy}—freeze the aligned \bcos backbone for linear probing and zero-shot transfer; faithful, well-localized explanations follow directly from \(W(\mathbf{x})\) (\ie, model summary), amortizing interpretability across tasks (\cref{sec:main_results}). 
}
   \label{fig:setup}
\end{figure*}

\section{\ours: ALign Once to Explain}
\label{sec:aloe}

In this section, we introduce \textbf{\ours} \textit{ (\textbf{AL}ign \textbf{O}nce to \textbf{E}xplain)}, a \emph{one-time, label-free feature-alignment} procedure that converts a ViT-based foundation encoder into a \bcos~\citeMain{boehle2024bcos} counterpart and aligns it to a frozen teacher via representation matching (\cref{fig:setup}).
We first describe the \emph{teacher–student} setup and the architecture-preserving transformation used to initialize the student (\cref{sec:aloe_setup}). Next, we detail what is aligned and where in the network, including ViT-specific tokens (e.g., \texttt{[CLS]} and registers) and intermediate layers used to guide alignment (\cref{sec:aloe_alignment}). Finally, we describe our setup (\cref{sec:aloe_train}) and show how the aligned, inherently interpretable backbone can be used as a drop-in component across downstream tasks, thereby amortizing the cost of interpretability (\cref{sec:aloe_eval}).

\subsection{Teacher--student setup}
\label{sec:aloe_setup}

\myparagraph{\bcos conversion.} Given a frozen \emph{teacher} encoder $\mathcal{T}$ (supervised, self-supervised, or vision–language), we construct a \bcos \emph{student} $\mathcal{S}$ by applying the architecture-preserving transformation recipe~\citeMain{arya2024b}~(see~\cref{sec:background_bcosification}), which preserves functional behavior where possible while requiring only minimal edits needed for inherent interpretability.

\myparagraph{Preserving special tokens.} Unlike CNNs, modern ViTs use special tokens such as \texttt{[CLS]} and register tokens (\eg, in DINOv3) that are crucial for performance. Given this, we keep these special-token pathways identical to those of the teacher so subsequent alignment can match tokens one-to-one and preserve the base model’s computational routing\footnote{SigLIP2~\citeMain{tschannen2025siglip} image encoders do not use \texttt{[CLS]} or register tokens.}. 

These steps yield a bias-free, 6-input channel \bcos student that architecturally mirrors the teacher closely and is ready for label-free alignment.

\subsection{Label-free alignment}
\label{sec:aloe_alignment}

After conversion (\cref{sec:aloe_setup}), we \emph{align} the \bcos student $\mathcal{S}$ to the frozen teacher $\mathcal{T}$ using unlabeled images and a cosine-similarity objective. 
Our aim is to make the aligned \bcos backbone a \emph{drop-in} substitute for the foundation teacher. This motivates (i) a \emph{global} guidance term to preserve the geometry of the final embedding space---critical for downstream tasks (classification, dense prediction, zero-shot transfer)---and (ii) \emph{token-level, depth-wise} guidance to preserve intermediate computations and improve optimization stability. Therefore, we propose a multi-layer objective that is applied across selected depths of the two models during training.

\myparagraph{Alignment objective.} 
For an input $\mathbf{x}$, let $E_\star(\mathbf{x})$ denote the model's last-layer image representation, and let $h_{\star,t}^{\ell}(\mathbf{x})$ be the token-level features at transformer block $\ell$ for token $t$.
We combine a global term with depth-wise layer guidance:

\begin{equation}
\label{eq:loss-total}
\mathcal{L} \;=\; \lambda_{\mathrm{g}}\,\mathcal{L}_{\mathrm{global}} \;+\; \lambda_{\mathrm{l}}\,\mathcal{L}_{\mathrm{layers}} \,,
\end{equation}

\begin{equation}
\label{eq:loss-global}
\mathcal{L}_{\mathrm{global}}
= \sum_{\mathbf{x}\in\mathcal{B}}\frac{1}{|\mathcal{B}|}
\Big(1 - \cos\big(E_{\mathcal{S}}(\mathbf{x}),\,E_{\mathcal{T}}(\mathbf{x})\big)\Big),
\end{equation}

\begin{equation}
\label{eq:loss-layers}
\begin{aligned}
\mathcal{L}_{\mathrm{layers}}
&= \sum_{\mathbf{x}\in\mathcal{B}}\frac{1}{|\mathcal{B}|}
   \sum_{\ell\in\mathcal{L}_{\mathrm{depth}}}
   \frac{1}{\big|\mathcal{T}_{\mathrm{tok}}^{\ell}(\mathbf{x})|} \\
&\qquad \times \sum_{t\in\mathcal{T}_{\mathrm{tok}}^{\ell}(\mathbf{x})}
   \big(1 - \cos\!\big(h^{\ell}_{\mathcal{S},t}(\mathbf{x}),\,h^{\ell}_{\mathcal{T},t}(\mathbf{x})\big)\big),
\end{aligned}
\end{equation}

\noindent where $\mathcal{T}_{\mathrm{tok}}^{\ell}(\mathbf{x})$ denotes the set of teacher tokens at depth $\ell$.

We use cosine as our default alignment loss because it is scale-invariant and robust across teacher sizes and pre-training paradigms. Feature scales vary widely in practice; cosine normalizes this mismatch and directly optimizes angular agreement, which aligns well with objectives used during pre-training~\citeMain{zhai2023sigmoid,simeoni2025dinov3}. We also experimented with compatible objectives (MSE, SigLIP, InfoNCE) and found cosine to be the most stable across pre-trained teachers; ablations are reported in~\cref{sec:main_ablations}. In contrast, MSE is sensitive to absolute scale, and contrastive variants (InfoNCE/SigLIP) introduce batch negatives that likely distort the teacher's local geometry. 

\myparagraph{Targets and depth-wise guidance.}
We supervise at three evenly spaced depths
$\mathcal{L}_{\mathrm{depth}}=\{\lfloor L/3 \rfloor,\,\lfloor 2L/3 \rfloor,\,L\}$ for a transformer of depth $L$ (\ie, $1/3$, $2/3$, and full depth). We align exactly the semantics-carrying tokens used by each teacher (\eg, \texttt{[CLS]}/registers for DINOv3; attention pooling for SigLIP2; \cref{tab:global_local_tokens}), ensuring one-to-one routing and preserving the teacher's computational pathways---crucial for faithful \bcos explanations.

\begin{table}[h]
\centering
\caption{\textbf{Tokens used for alignment (\cref{sec:aloe_alignment}) for each model.}}
\footnotesize
\begin{tabular}{p{1.6cm} p{2.2cm} p{3cm}}
\textbf{Model} & \textbf{Global Feature} & \textbf{Layer-wise Features} \\
\midrule
Supervised~\citeMain{dosovitskiy2021an} 
& Final \texttt{[CLS]} 
& Image tokens, \texttt{[CLS]} \\
\midrule
DINOv3~\citeMain{simeoni2025dinov3} 
& Final \texttt{[CLS]} 
& Image tokens, \texttt{[CLS]}, 

register tokens \\

\midrule
SigLIP2~\citeMain{tschannen2025siglip} 
& Attention-pooled final embedding 
& Image tokens \\
\bottomrule
\end{tabular}
\label{tab:global_local_tokens}
\end{table}

\noindent 
 This multi-scale supervision improves optimization stability and final alignment. Because the student $\mathcal{S}$ mirrors the teacher $\mathcal{T}$ in width, no projection heads are required. In practice, $(\lambda_{\mathrm{g}},\lambda_{\mathrm{l}})=(1,1)$ works well across all models. We ablate the design choices---including the alignment objective and feature depths---to select the final configuration empirically (\cref{sec:main_ablations}).

\subsection{Implementation and training setup}
\label{sec:aloe_train}

We now describe the implementation and training setup for feature alignment.

\myparagraph{Datasets.}
We use the unlabeled, web-scale image sets CC3M~\citeMain{sharma2018conceptual}, CC12M~\citeMain{changpinyo2021conceptual}, and YFCC15M~\citeMain{cui2022democratizing} for alignment. The input resolution follows the teacher's default (typically $224{\times}224$). For our main results (\cref{tab:lp_all}), we report numbers for models trained on YFCC15M.

\myparagraph{Teachers and students.}
Teachers $\mathcal{T}$ are strong ViT-based foundation encoders spanning three pre-training paradigms:
(i) \textit{Supervised} ViT-B/16~\citeMain{dosovitskiy2021an};
(ii) \textit{DINOv3 (self-supervised)} ViT-S/16, ViT-B/16, and ViT-L/16~\citeMain{simeoni2025dinov3};
(iii) \textit{SigLIP2 (contrastive vision-language)} ViT-B/16, ViT-L/16, ViT-so400m/16, and ViT-so400m/14@384~\citeMain{tschannen2025siglip}.
B-cos students $\mathcal{S}$ are constructed using the architecture-preserving conversion recipe (\cref{sec:background_bcosification}) and initialized to mirror $\mathcal{T}$ in width, depth, and token layout, enabling one-to-one alignment (\cref{sec:aloe_alignment}). As shown in \cref{fig:setup}, the linear patch embedding is replaced by a B-cos convolutional stem with the same overall stride, so the student's token grid matches the teacher's (see \cref{supp:sec:impl} for SigLIP2 so400m/14@384).

\myparagraph{Augmentations.}
We apply random resized crops and horizontal flips as augmentations to prevent overfitting while preserving $\mathcal{T}$--$\mathcal{S}$ feature geometry during alignment.

\myparagraph{Optimization.}
We freeze $\mathcal{T}$ and optimize $\mathcal{S}$ with AdamW~\citeMain{loshchilov2017decoupled} and a cosine learning-rate schedule, with mixed precision enabled. We follow an early-stopping criterion to train until the loss on a held-out validation set stops decreasing; we fix $B{=}2$, keep additive biases at zero, and use no explicit weight normalization (cf.\ \cref{sec:background_bcosification}). Gradient norm clipping of 1.0 (with no weight decay) is used to stabilize large models. We sweep learning rates over \texttt{3e-3}, \texttt{1e-3}, and \texttt{5e-4} and select the model with the lowest alignment loss on the held-out split (a randomly sampled 30k subset of images from the training set). We use a batch size of 1024 for all experiments.
All unmarked results use checkpoints trained with the updated B-cos implementation and without the temporary weight-norm stabilization term that earlier checkpoints used to counter a potential training instability; \prevreg marks results from earlier checkpoints, including the ablations (see \cref{supp:sec:prevreg}). B-cosification~\citeMain{arya2024b} baselines use the original public implementation.

\myparagraph{Loss arguments.}
Unless specified: $(\lambda_{\mathrm{g}},\lambda_{\mathrm{l}})=(1,1)$; cosine-similarity alignment loss; and depth supervision at $\{\lfloor L/3 \rfloor,\lfloor 2L/3 \rfloor,L\}$. We ablate alignment objectives, feature depths, model sizes, and dataset scales in \cref{sec:main_ablations}.

\subsection{Evaluation protocols and downstream usage}
\label{sec:aloe_eval}

\emph{Train minimally, explain everywhere.}  Once aligned, the \bcos backbone is used as a drop-in encoder across tasks; explanations are model-inherent (derived from $\mathbf{W}(\mathbf{x})$~\citeMain{boehle2024bcos}) and require no task-specific interpretability tuning.

\myparagraph{Linear probing (LP).}
We evaluate the aligned representation quality of $\mathcal{S}$ by training a linear head on frozen features, reporting top-1 accuracy on the validation sets of 10 downstream classification datasets (including IN1k) following the standard protocol~\citeMain{oquab2024dinov} (see~\cref{sec:main_results}). To speed up evaluation, we do \emph{not} apply augmentations during probing.

\myparagraph{$k$-NN evaluation.}
To assess feature quality without additional training, we use a weighted $k$-NN classifier on frozen embeddings. We pre-compute features on the training split and use $k{=}20$, a robust choice across datasets~\citeMain{caron2021emerging} and report accuracy (\%) on the validation split(s).

\myparagraph{Zero-shot evaluation.}
For the contrastive vision–language setting (SigLIP2), we use the ALOE-aligned \bcos image encoder as a \textit{drop-in} replacement while keeping the \emph{original} text encoder unchanged. We use the 80 ImageNet prompt templates of~\citeMain{radford2021learning}, averaging text embeddings per class; temperature and normalization follow the default settings. We evaluate zero-shot classification on standard benchmarks~\citeMain{radford2021learning, zhai2023sigmoid} and report accuracy (\%).

\myparagraph{Dense prediction.}
We evaluate dense prediction by training a linear depth head on frozen features and reporting NYUv2~\citeMain{silberman2012indoor} depth metrics.

\myparagraph{Evaluating explanations.}
We evaluate explanation quality using the Grid Pointing Game (GridPG)~\citeMain{boehle2021convolutional,zhang2018top} for localization and pixel-deletion for faithfulness. For conventional models, we compare against several post-hoc methods (AttnLRP~\citeMain{achtibat2024attnlrp}, Integrated Gradients~\citeMain{sundararajan2017axiomatic}, LeGrad~\citeMain{bousselham2024legrad}, CheferCAM~\citeMain{chefer2021transformer}, and LIME~\citeMain{ribeiro2016should}); we also report results on a human preference study. Implementation details and additional results are provided in the appendix \cref{supp:sec:impl} and \cref{sec:supp_interp}.

\section{Results}
\label{sec:results}

In \cref{sec:main_results}, we compare \ours to baselines on the settings and metrics described in \cref{sec:aloe_eval}. Detailed ablations on alignment objectives, feature depths, model scales, and dataset size are provided in \cref{sec:main_ablations}.

\begin{table*}[t!]
\centering
\footnotesize
\caption{\textbf{Linear-probe accuracy on base architectures (ViT-B/16).} \ours substantially outperforms vanilla B-cosification while remaining competitive with the original foundation models. Teachers are shown in gray; best interpretable model in \textbf{bold}. \textit{Note: For DINOv3 B-cosification we follow the supervised fine-tuning protocol~\cite{arya2024b} since the original work did not have a self-supervised baseline. All B-cosification baselines use the original public implementation (\cref{supp:sec:prevreg}).}}
\label{tab:lp_all}
\setlength{\tabcolsep}{4.5pt} 
\begin{tabular}{@{} R c R R R R R R R R R R R@{}}
\toprule
\multicolumn{1}{R}{\textbf{Feature}} &
\multicolumn{1}{c}{\textbf{Inh. Inter.}} &
\multicolumn{1}{c}{\textbf{IN1k}} &
\multicolumn{1}{c}{\textbf{Cal101}} &
\multicolumn{1}{c}{\textbf{Flowers}} &
\multicolumn{1}{c}{\textbf{Food}} &
\multicolumn{1}{c}{\textbf{Aircr}} &
\multicolumn{1}{c}{\textbf{DTD}} &
\multicolumn{1}{c}{\textbf{Cars}} &
\multicolumn{1}{c}{\textbf{SUN}} &
\multicolumn{1}{c}{\textbf{C10}} &
\multicolumn{1}{c}{\textbf{C100}} &
\multicolumn{1}{c}{\textbf{Avg.}} \\
\midrule

\textcolor{mygray}{Sup.~\citeMain{dosovitskiy2021an}} & \xmark & \textcolor{mygray}{81.19} & \textcolor{mygray}{97.43} & \textcolor{mygray}{99.33} & \textcolor{mygray}{86.22} & \textcolor{mygray}{41.31} & \textcolor{mygray}{74.44} & \textcolor{mygray}{57.39} & \textcolor{mygray}{74.19} & \textcolor{mygray}{97.07} & \textcolor{mygray}{86.56} & \textcolor{mygray}{79.51} \\
\addlinespace[2pt]
B-cosif.~\citeMain{arya2024b} & \cmark  & 77.65 & 96.35 & 95.83   & 78.18 & 37.07 & 71.15 & 46.06 & 66.83 & 95.00 & 81.43 & 74.56 \\
\rowcolor{hirow}
\ours (ours) & \cmark  & \textbf{81.30} & \textbf{97.32} & \textbf{99.44} & \textbf{86.73} & \textbf{43.96} & \textbf{75.55} & \textbf{61.59} & \textbf{74.99} & \textbf{97.53} & \textbf{87.67} & \textbf{80.61} \\
\addlinespace[2pt]
& & \textcolor{teal}{+3.65} & \textcolor{teal}{+0.97} & \textcolor{teal}{+3.61} & \textcolor{teal}{+8.55} & \textcolor{teal}{+6.89} & \textcolor{teal}{+4.40} & \textcolor{teal}{+15.53} & \textcolor{teal}{+8.16} & \textcolor{teal}{+2.53} & \textcolor{teal}{+6.24} & \textcolor{teal}{+6.05} \\

\midrule
\setrowstyle{\color{mygray}}
\textcolor{mygray}{SigLIP2}~\citeMain{tschannen2025siglip} & \xmark & 84.21 & 98.88 & 99.21 & 94.44 & 75.48 & 85.74 & 95.42 & 81.70 & 96.95 & 84.91 & 89.69 \setrowstyle{} \\
\addlinespace[2pt]
B-cosif.~\citeMain{arya2024b} & \cmark  & 75.05 & 97.52 & 94.14 & 83.74 & 45.04 & 79.18 & 73.70 & 75.68 & 94.44 & 81.11 & 79.96 \\
\rowcolor{hirow}
\ours (ours) & \cmark & \textbf{83.87} & \textbf{98.32} & \textbf{98.84} & \textbf{94.20} & \textbf{72.47} & \textbf{83.03} & \textbf{94.31} & \textbf{81.52} & \textbf{96.71} & \textbf{84.53} & \textbf{88.78} \\
\addlinespace[2pt]
& & \textcolor{teal}{+8.82} & \textcolor{teal}{+0.80} & \textcolor{teal}{+4.70} & \textcolor{teal}{+10.46} & \textcolor{teal}{+27.43} & \textcolor{teal}{+3.85} & \textcolor{teal}{+20.61} & \textcolor{teal}{+5.84} & \textcolor{teal}{+2.27} & \textcolor{teal}{+3.42} & \textcolor{teal}{+8.82} \\

\midrule
\setrowstyle{\color{mygray}}
\textcolor{mygray}{DINOv3}~\citeMain{simeoni2025dinov3} & \xmark & 84.36 & 98.88 & 99.55 & 94.12 & 80.40 & 83.98 & 94.51 & 78.74 & 98.22 & 89.24 & 90.20 \setrowstyle{} \\
\addlinespace[2pt]
B-cosif.~\citeMain{arya2024b} & \cmark & 73.64 & 95.18 & 82.68 & 68.24 & 41.67 & 68.02 & 50.90 & 58.25 & 92.04 & 76.44 & 70.71 \\
\rowcolor{hirow}
\ours (ours) & \cmark & \textbf{84.14} & \textbf{98.77} & \textbf{99.55} & \textbf{93.86} & \textbf{80.19} & \textbf{83.42} & \textbf{94.41} & \textbf{78.23} & \textbf{98.01} & \textbf{89.35} & \textbf{89.99} \\
\addlinespace[2pt]
& & \textcolor{teal}{+10.50} & \textcolor{teal}{+3.59} & \textcolor{teal}{+16.87} & \textcolor{teal}{+25.62} & \textcolor{teal}{+38.52} & \textcolor{teal}{+15.40} & \textcolor{teal}{+43.51} & \textcolor{teal}{+19.98} & \textcolor{teal}{+5.97} & \textcolor{teal}{+12.91} & \textcolor{teal}{+19.28} \\
\bottomrule
\end{tabular}
\end{table*}

\subsection{Main results}
\label{sec:main_results}

\myparagraph{Linear evaluation of frozen features.} We compare \ours to (i) the original foundational models and (ii) B-cosification~\citeMain{arya2024b}. All use the same linear protocol and input resolution (see ~\cref{sec:aloe_eval}).

In Table~\ref{tab:lp_all}, we report linear probing results on frozen features on ten datasets (\citeMain{chen2021empirical}).
\ours improves markedly over B-cosification across ten datasets. For ViT-B/16, average gains vs.\ B-cosification are: \textbf{+6.05} p.p.\ (from \textbf{$74.56\%$} $\rightarrow$ \textbf{$80.61\%$}) for fully supervised; \textbf{+8.82} p.p.\ (\textbf{$79.96\%$} $\rightarrow$ \textbf{$88.78\%$}) for SigLIP2; and \textbf{+19.28} p.p.\ (\textbf{$70.71\%$} $\rightarrow$ \textbf{$89.99\%$}) for DINOv3. \ours also closely matches its teachers (e.g., SigLIP2 average $88.78\%$ vs.\ teacher $89.69\%$; DINOv3 $89.99\%$ vs.\ $90.20\%$). On ImageNet-1k linear probing, the SigLIP2-aligned B-ViT-B/16 reaches \textbf{83.87\%}; larger SigLIP2-aligned variants reach \textbf{87.08\%} and \textbf{87.83\%} for B-ViT-L/16 and B-ViT-so400m/16 (\cref{sec:supp_downstream}).

\begin{table}[t!]
  \centering
  \footnotesize
  
  \begin{subtable}[b]{0.48\linewidth}
    \centering
    \caption{$k$-NN Accuracy}
    \label{tab:knn_dinov3_b16}
    \begin{tabular}{@{}lcc@{}}
      \toprule
      \textbf{Method} & \textbf{IN1k (\%)} & \textbf{Avg (\%)} \\
      \midrule
      \textcolor{mygray}{DINOv3}~\citeMain{simeoni2025dinov3}        & \textcolor{mygray}{82.28} & \textcolor{mygray}{83.71}  \\
      B-cosif.~\citeMain{arya2024b} & 71.03 & 50.78 \\
      \ours & \textbf{81.56} & \textbf{83.38} \\
      \bottomrule
    \end{tabular}
  \end{subtable}\hfill
  \begin{subtable}[b]{0.43\linewidth}
    \centering
    \caption{Zero-shot Accuracy}
    \label{tab:zs_siglip2_b16}
    \begin{tabular}{@{}lc@{}}
      \toprule
      \textbf{Method} & \textbf{@1 (\%)} \\
      \midrule
      \textcolor{mygray}{SigLIP2}~\citeMain{tschannen2025siglip}        & \textcolor{mygray}{78.07} \\
      B-cosif.~\citeMain{arya2024b} & 58.66 \\
      \ours             & \textbf{77.17} \\
      \bottomrule
    \end{tabular}
  \end{subtable}
  
  \vspace{0.5em}
  
  \caption{\textbf{Downstream and Zero-shot Evaluations.} \textbf{(a)} $k$-NN results ($k{=}20$). \textbf{(b)} Zero-shot ImageNet top-1 accuracy (SigLIP2 ViT-B/16, 80-prompt template). In both evaluation settings, ALOE substantially outperforms vanilla B-cosification and remains highly competitive with the original teacher models. }
  \label{tab:combined_evals}
\end{table}

Similar trends hold for $k$-NN evaluation. In Table~\ref{tab:knn_dinov3_b16} we report accuracy on IN1k~\citeMain{imagenet} for \ours vs B-cosification; we get \textbf{+10.53}~p.p.\ on IN1k and \textbf{+32.60}~p.p.\ averaged across datasets (see \cref{sec:supp_downstream} for additional results).

\myparagraph{Zero-shot classification.} \ours preserves strong zero-shot transfer and outperforms B-cosification. On ImageNet-1k (zero-shot@1) with SigLIP2 ViT-B/16, \ours achieves \textbf{77.17}\% vs.\ B-cosification \textbf{58.66}\%, close to the SigLIP2 teacher at \textbf{78.07}\% (Tab.~\ref{tab:zs_siglip2_b16}).

\myparagraph{Dense prediction.}
Beyond image-level recognition, \ours also preserves spatially structured features useful for dense prediction. On monocular depth estimation with a linear probe (ViT-B/16), \ours substantially outperforms B-cosification on both relative and absolute metrics, improving relative $\delta_1$ from \textbf{0.83} to \textbf{0.93} and reducing RMSE from \textbf{0.46} to \textbf{0.33}, while approaching the DINOv3 teacher (\textbf{0.95}/\textbf{0.28}); see \cref{sec:supp_downstream} for more dense results, including surface normals and multiview correspondence.

\begin{figure}[ht!]
  \begin{center}
    \centering
        \includegraphics[width=0.95\linewidth]{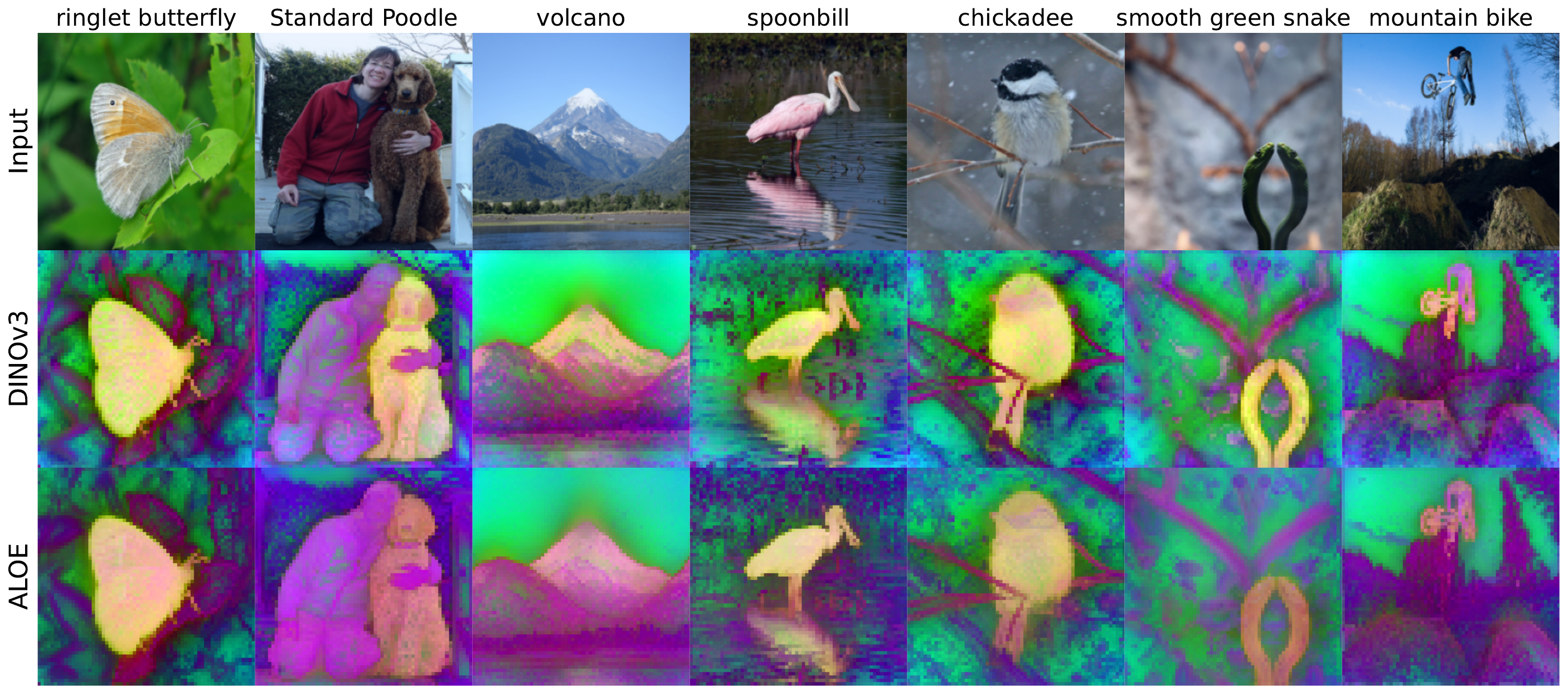}
\end{center}
    
 \caption{\textbf{PCA Visualization.} DINOv3 (\textbf{row 2}) vs. ALOE (\textbf{row 3}) last layer features visualized with 3 principal components and mapped to joint RGB space, preserves global feature geometry that indicates aligned semantics while being inherently interpretable.}

   \label{fig:pca}
\end{figure}

\myparagraph{PCA Visualization.} To visualize alignment quality, we project last-layer features to RGB using the first three principal components with shared scaling per teacher–student pair. As shown in Fig.~\ref{fig:pca}, ALOE preserves the teacher’s global feature geometry (e.g., DINOv3 vs.\ ALOE), indicating aligned semantics while being inherently interpretable.

\begin{figure}[t!]
\begin{center}
    \centering
        \includegraphics[width=\linewidth]{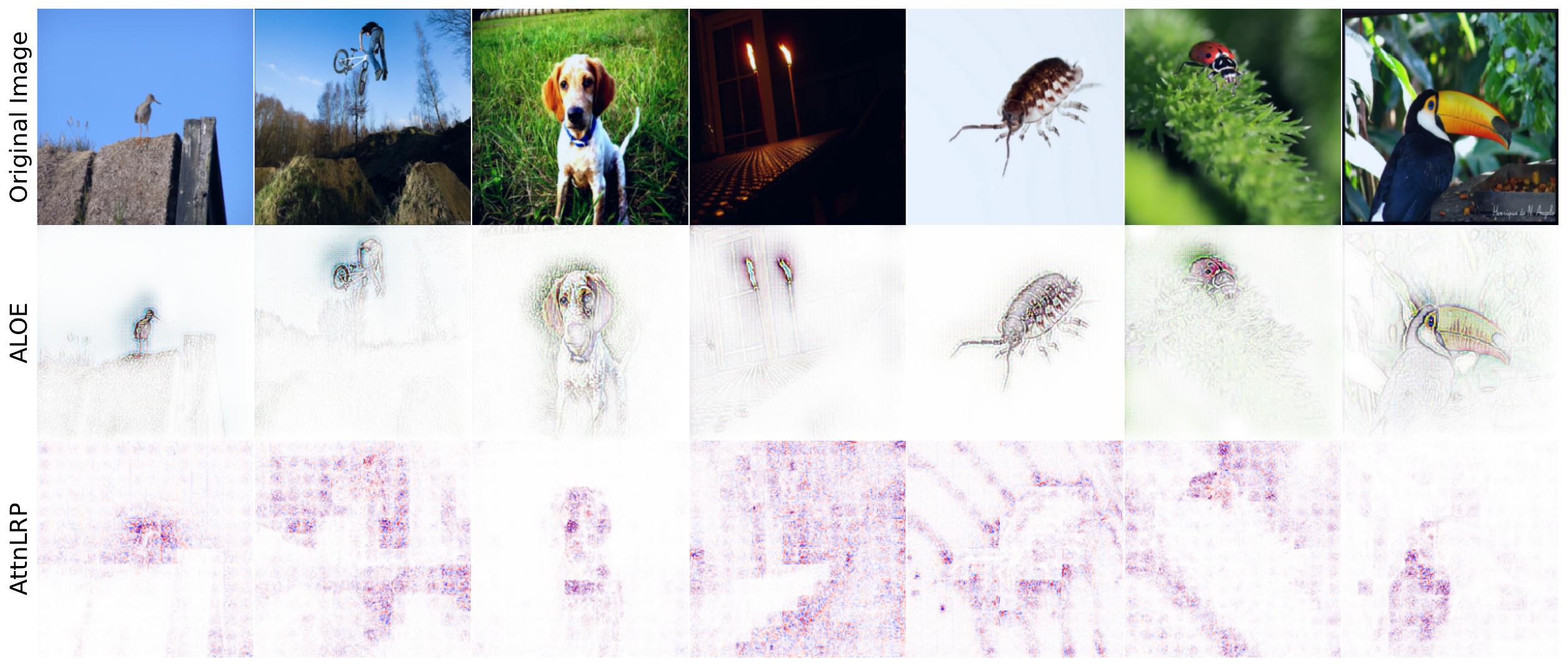}
\end{center}
    
 \caption{\textbf{Explanation quality.} Comparison between model inherent (\bcos: W(x)~\citeMain{boehle2024bcos}) explanations after \ours vs a popular strong method for visualizing vision transformers, AttnLRP~\citeMain{achtibat2024attnlrp}.}

   \label{fig:aloe_vs_attnlrp}
\end{figure}

\myparagraph{Interpretability.} We evaluate explanation quality for: (a) \emph{model-inherent} \bcos attributions $\mathbf{W}(\mathbf{x})\mathbf{x}$ from our ALOE-aligned models, and (b) post-hoc attributions on conventional models (AttnLRP~\citeMain{achtibat2024attnlrp} here). Following \citeMain{boehle2024bcos}, we report GridPG (higher is better), and defer additional faithfulness metrics, human evaluation, and post-hoc comparisons to \cref{sec:supp_interp}. Across supervised, SigLIP2, and DINOv3 teachers, \ours yields higher localization scores than AttnLRP (see~\cref{fig:teaser}, right). For example, for the SigLIP2 model, \ours attains \textbf{84.4}\% vs.\ teacher (AttnLRP) \textbf{55.5}\%, and is competitive with B-cosification.

In \cref{fig:aloe_vs_attnlrp}, we show qualitative explanation maps, \ours (\bcos: $W(x)$) and teacher (AttnLRP) for a DINOv3 ViT-B/16 model. The explanations for \ours (\textbf{row 2}) appear to align with class-discriminant patterns in the input and resemble the class objects. This is a result of alignment pressure during optimization (\cf.~\citeMain{boehle2022bcos}). In contrast, the AttnLRP explanations are sparse and noisy (\textbf{row 3}).

\begin{figure}[h!]
  \centering
    \begin{subfigure}{\linewidth}
    \includegraphics[width=\linewidth]{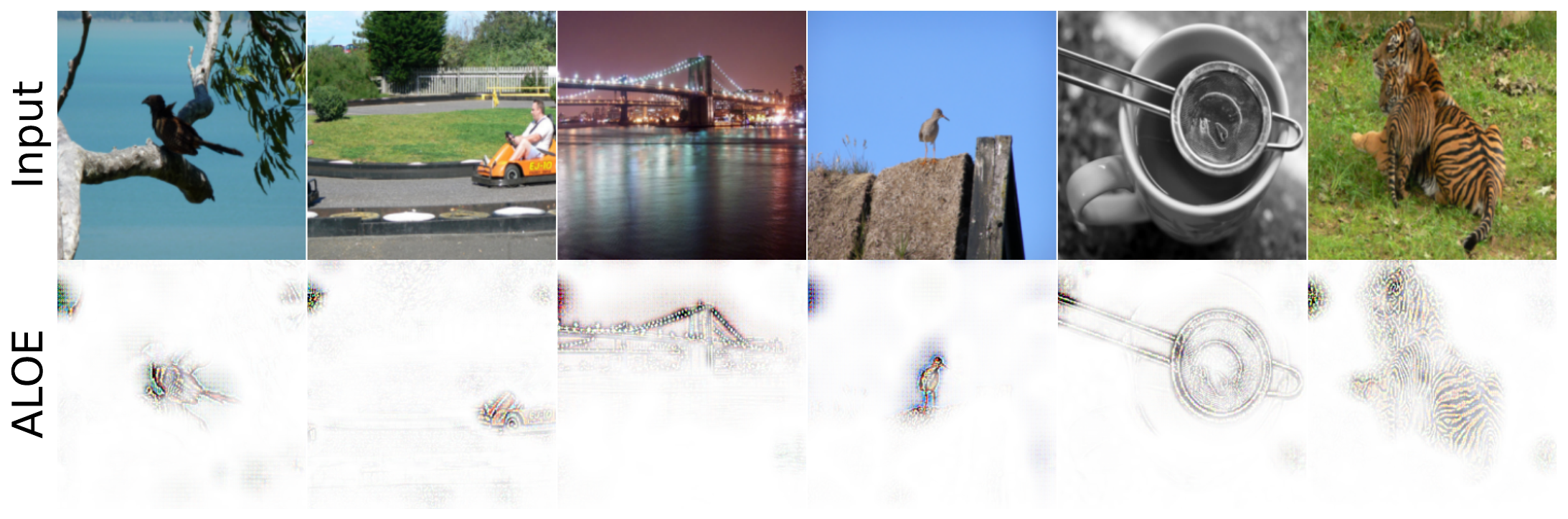}
  \end{subfigure}
  
 \caption{\textbf{VLM Zero-Shot Explanations.} \ours aligned \bcos models (SigLIP2 teacher) yield zero-shot model-inherent explanations for VLMs, where the input prompt to the text encoder is ``this is a photo of \{class name\}''. The explanations are visually well aligned with class-specific features.
 }
   \label{fig:siglip2_zeroshot}
\end{figure}

\begin{figure}[h!]
  \begin{center}
    \centering
        \includegraphics[width=0.8\linewidth]{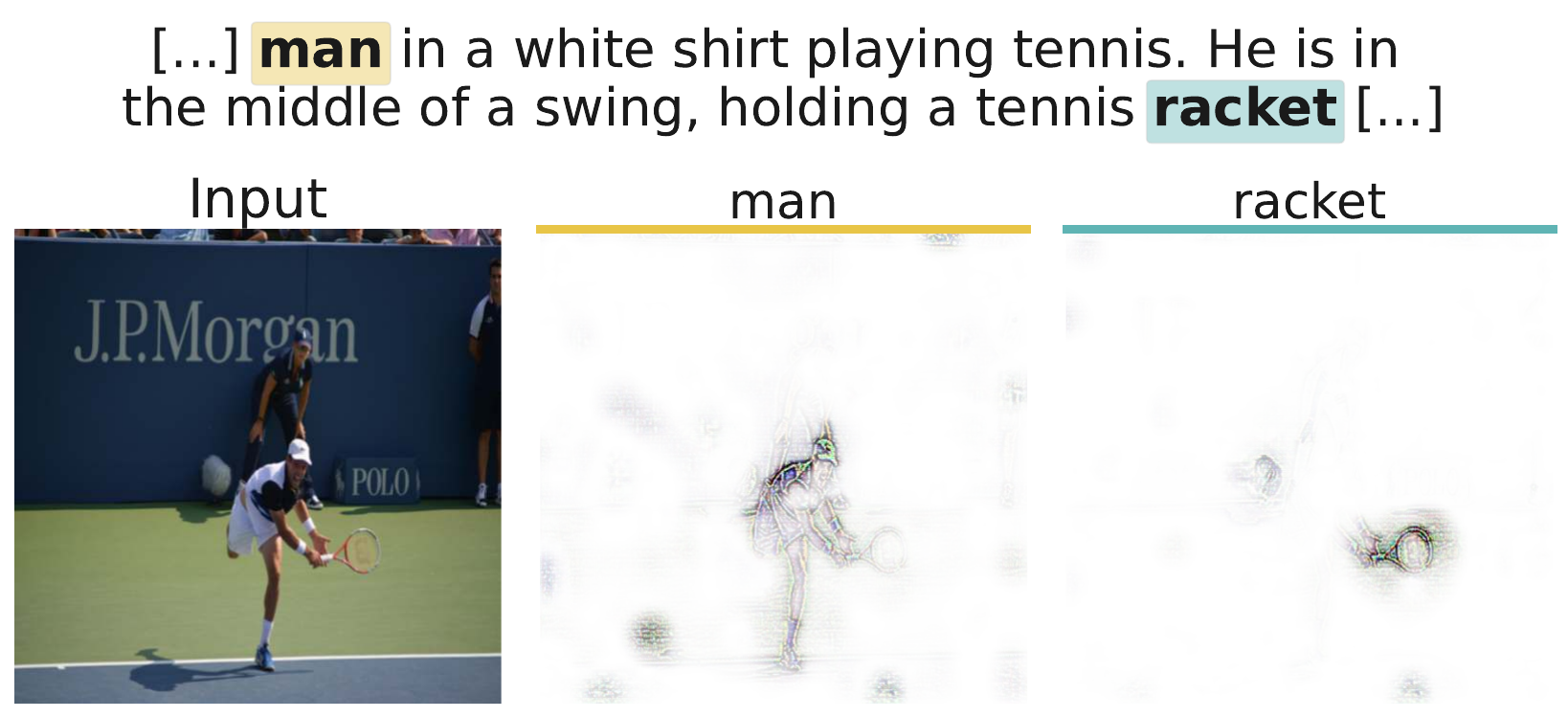}
\end{center}
  \caption{\textbf{Token-level Visual Grounding with \ours\prevreg.} By propagating relevance through the LLaVA-MORE \cite{cocchi2025llava} Gemma-2-9B \cite{gemmateam2024gemma2improvingopen} backbone (using AttnLRP) and extracting visual explanations from our \bcos SigLIP2 encoder, we achieve fine-grained localization for individual generated tokens. The relevance maps highlight well their corresponding semantic regions in the input image. Further examples and failure cases are shown in \cref{supp:sec:qualitative}. \prevregnote
  }
  \label{fig:llava_grounding}
\end{figure}

In \cref{fig:siglip2_zeroshot}, we show zero-shot explanations for \ours ViT-B/16 models aligned with SigLIP2. This lets us use the model as an explainable visual backbone combined with the corresponding text encoder. We use the prompt ``this is a photo of \{class name\}'' to compute similarities between text and visual features. The resulting explanations are highly localized and class-specific, yielding inherently explainable vision--language models.

\myparagraph{Multimodal LLMs.} More broadly, our aligned SigLIP2 backbones can also serve as interpretable visual encoders within multimodal large language models (MLLMs). In preliminary experiments with a LLaVA-style Gemma-2-9B backbone~\citeMain{gemmateam2024gemma2improvingopen}, they enable fine-grained visual grounding for individual generated tokens (see \cref{fig:llava_grounding}), suggesting that ALOE can extend beyond pure vision tasks. Since the current setup still relies on AttnLRP~\citeMain{achtibat2024attnlrp} to propagate relevance through the language model, we defer a fuller treatment of end-to-end inherently interpretable MLLMs to future work.

\subsection{Ablations}
\label{sec:main_ablations}

We ablate key design choices: \textit{multi-layer feature alignment}, \emph{model scale}, \emph{data efficiency} and \emph{alignment objectives}. For all ablations, we train for 40 epochs. These ablations use earlier-implementation runs (\prevreg, \cref{supp:sec:prevreg}): we first compared alignment objectives using only the global term, and then used the selected objective to ablate feature depth.

\begin{table}[h!]
\centering
\caption{\textbf{Feature depth ablation\prevreg.} Average LP accuracy over 10 datasets for SigLIP2 ViT-B/16 (cosine loss). \prevregnote}
\footnotesize
\setlength{\tabcolsep}{3.5pt}
\begin{tabular}{@{}lcccc@{}}
\toprule
Pool & $+L$ & $+\{2/3,\,L\}$ & $+\{1/3,\,2/3,\,L\}$ & $+$All \\
\midrule
80.86 & 83.06 & 88.66 & \textbf{88.79} & 88.41 \\
\bottomrule
\end{tabular}
\label{tab:where_loss_cols}
\end{table}

\myparagraph{Feature Depth.}
We ablate loss placement across depth(s): \textit{global}-only; \textit{global}$+L$; \textit{global}$+\{\lfloor 2L/3\rfloor,\,L\}$; \textit{global}$+\{\lfloor L/3\rfloor,\,\lfloor 2L/3\rfloor,\,L\}$; and \textit{global}$+$all. For SigLIP2 ViT-B/16 (Tab.~\ref{tab:where_loss_cols}), average LP accuracy improves with increasing depth and peaks at $\{\lfloor L/3\rfloor,\lfloor 2L/3\rfloor,L\}$; supervising all layers slightly degrades performance (88.41 vs.\ 88.79). We thus use 3 evenly spaced depths.

\myparagraph{Model scale.}
We see consistent gains in accuracy with increasing model size. For this we align DINOv3 and SigLIP2 teachers (and their \bcos students). Larger models close the gap to the teacher foundation model (see~\cref{fig:aloe_vs_bcosification}).

\begin{figure}[h!]
  \centering
      \includegraphics[width=0.9\linewidth]{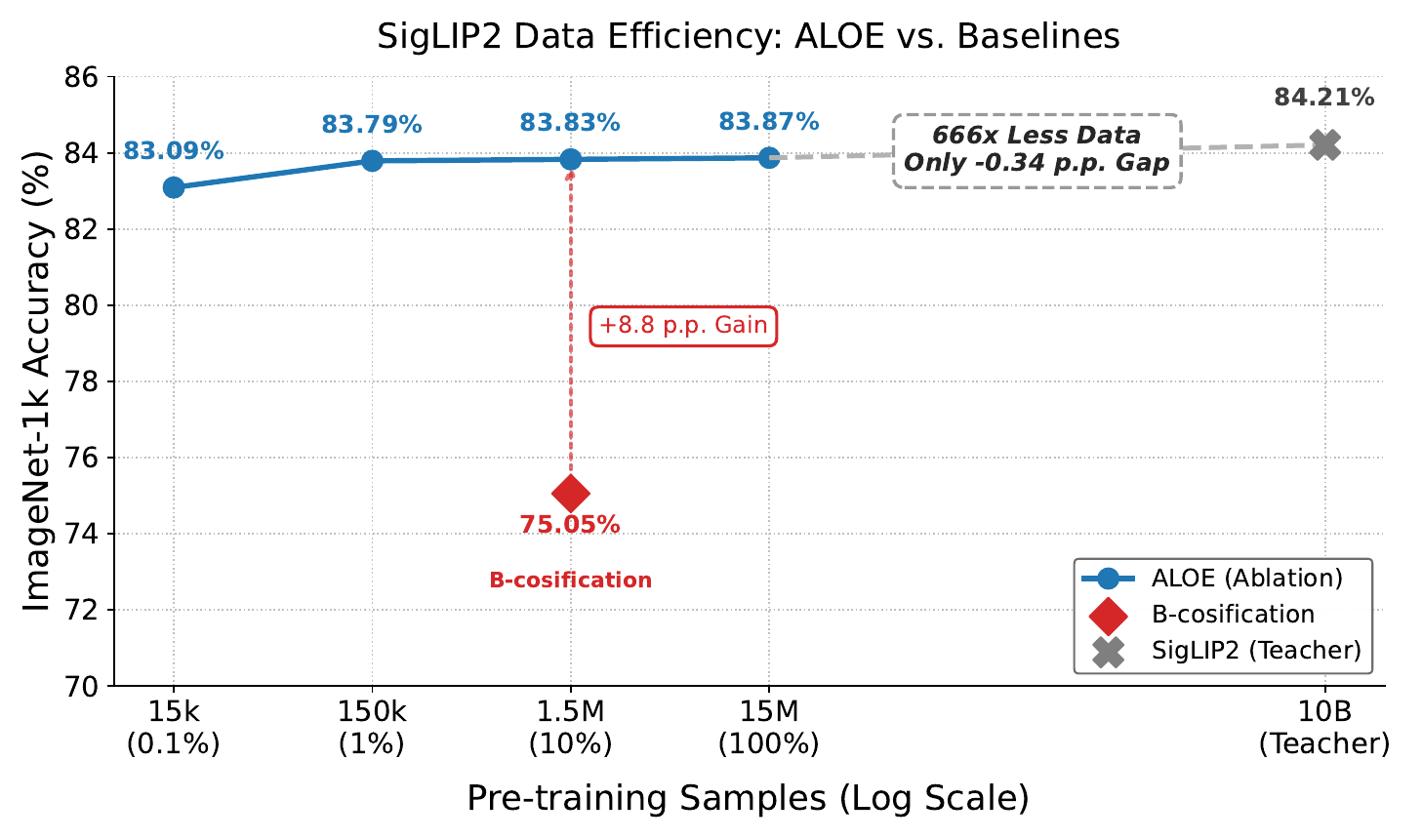}
  \caption{\textbf{Data efficiency of \ours.}
  ImageNet top-1 linear-probe accuracy for \ours aligned SigLIP2 (ViT-B/16) across varying amounts of YFCC15M data (0.1\%--100\%). Across these settings, including the full-data distillation, \ours remains nearly flat (83.09\%--83.87\%). Already at 1\% (150k images), it reaches 83.79\%, within 0.42 p.p. of the SigLIP2 teacher (84.21\%) and 8.74 p.p. above the B-cosification reference. The 100\% setting reaches 83.87\%, leaving only a 0.34 p.p. gap to the teacher.}
  \label{fig:aloe_siglip_data_scaling}
\end{figure}

\myparagraph{Data efficiency.}
Using only unlabeled YFCC15M~\citeMain{cui2022democratizing}, \ours maintains nearly flat ImageNet-1k top-1 linear-probe accuracy when the alignment data shrinks from 100\% to 0.1\%, staying within 0.78 p.p.; see \cref{fig:aloe_siglip_data_scaling} for SigLIP2. Already at 1\% of YFCC15M (150k images), the model is within 0.42 p.p. of the SigLIP2 teacher, highlighting the strong data efficiency of our one-time feature-alignment procedure. 

\myparagraph{Alignment objectives.}
Under identical settings with only the global alignment term (\cref{tab:loss_ablation_in1k_lp}), \textit{cosine} and \textit{SigLIP} are most consistent across models, \textit{MSE} and \textit{InfoNCE} showed inconsistencies. Given the ease of application and simplicity we adopt \textbf{cosine} by default.

\section{Conclusion}
\label{sec:conclusion}

We presented \textbf{\ours} (ALign Once to Explain), a simple yet effective framework that converts ViT-based foundational encoders into inherently interpretable \bcos variants. By preserving special tokens and aligning both global embeddings and token-level features at a few depths with a cosine objective, \ours yields backbones that achieve competitive performance as compared to conventional foundation models across a diverse set of downstream tasks, provide faithful, well-localized explanations by design, and scale across pre-training paradigms and model sizes with modest amounts of unlabeled data.

Compared to B-cosification~\citeMain{arya2024b}, \ours removes the need for supervised fine-tuning, addresses the ViT performance gap, and preserves contrastive zero-shot utility—offering a resource-efficient path to interpretable backbones at foundation scale. We show empirically that our proposed recipe can already recover most of the teacher’s generalization from a small unlabeled alignment set, even with only 150k images in the SigLIP2 data-scaling ablation. Taken together, \ours offers a practical path to scalable, trustworthy vision backbones: \textit{align once, reuse broadly, and amortize the cost of interpretability.}






\section{Citing this work}
\label{sec:citation}

{\color{black}
\noindent This is the arXiv version of a paper
accepted at CVPR 2026. Please cite the CVPR version~\cite{Maser_2026_CVPR}. 
BibTeX entry given below:
}

\vspace{0.6em}
\noindent\textbf{CVPR version:}
\vspace{0.2em}
\hrule
\vspace{0.4em}
{\footnotesize
\begin{verbatim}
@inproceedings{maser2026aloe,
  title     = {Align Once to Explain: Feature
               Alignment for Scalable
               B-cosification of Foundational
               Vision Transformers},
  author    = {Maser, Raphael and Gairola,
               Siddhartha and Rao, Sukrut and
               Schiele, Bernt},
  booktitle = {Proceedings of the IEEE/CVF
               Conference on Computer Vision
               and Pattern Recognition (CVPR)},
  year      = {2026}
}
\end{verbatim}
}
\hrule


\section*{Acknowledgements}

Computations were performed on the HPC systems at the Max Planck Computing and Data Facility (MPCDF). We thank Bart Pogodzinski and Wolfgang Boettcher for valuable feedback on the manuscript.

{
    \small
    \bibliographystyle{ieeenat_fullname}
    \bibliography{main}
}

\clearpage
\appendix

\renewcommand\thesection{\Alph{section}}
\numberwithin{equation}{section}
\numberwithin{figure}{section}
\numberwithin{table}{section}
\renewcommand{\thefigure}{\thesection\arabic{figure}}
\renewcommand{\thetable}{\thesection\arabic{table}}
\crefname{appendix}{Sec.}{Secs.}

{\onecolumn 
{\begin{center}
\Large\bf
\phantom{skip}\\[.25em]
{ALign Once to Explain~\texorpdfstring{\emoji{sec/static/aloe4.png}}{[aloe]}: Feature Alignment for Scalable B-cosification of Foundational Vision Transformers}\\[1em]
\large
Appendix
\end{center}
}

\newcommand{\additem}[2]{%
  \item[\textbf{(\ref{#1})}] \textbf{#2} \dotfill\makebox{\textbf{\pageref{#1}}}
}

\newcommand{\addsubitem}[2]{%
  \textbf{(\ref{#1})}\hspace{1em} #2\\[.1em]
}

\newcommand{\adddescription}[1]{%
  \par\noindent #1\par\vspace{1em}%
}

In this appendix to our work on scalable B-cosification of foundational Vision Transformers, we provide:
\begin{enumerate}[label={({\arabic*})}, topsep=.75em, itemsep=.25em]
  \additem{supp:sec:impl}{Implementation Details}
  \adddescription{Model checkpoints, datasets, evaluation protocols, and metrics.}

  \additem{supp:sec:prevreg}{Earlier Checkpoints and Dagger Notation}
  \adddescription{A note on the updated B-cos implementation, the B-cosification baselines, updated arXiv checkpoints, and dagger-marked results.}

  \additem{supp:sec:quantitative}{Additional Quantitative Results}
  \adddescription{Downstream performance (LP/$k$-NN, zero-shot, dense prediction), data scaling experiments, alignment objective analysis and interpretability evaluations.}

  \additem{supp:sec:qualitative}{Additional Qualitative Results}
  \adddescription{Zero-shot explanations, comparisons to popular post-hoc methods, and depth estimation visualizations.}
\end{enumerate}

\setlength{\parskip}{.5em}
\clearpage
\section{Implementation Details}
\label{supp:sec:impl}

In this section we provide additional implementation details that complement the main paper (\cref{sec:aloe_eval}), including details about the teacher checkpoints, datasets used for downstream evaluations, and finally the attribution methods along with the interpretability metrics we use.

\myparagraph{Teacher checkpoints.}
For our feature-alignment step (\cref{sec:aloe_alignment}) with pre-trained vision foundational models (\cite{dosovitskiy2021an,simeoni2025dinov3,tschannen2025siglip}), we align to frozen teacher encoders of publicly available checkpoints and keep the associated text encoders for vision-language models unchanged (for SigLIP2~\cite{tschannen2025siglip} zero-shot). Table~\ref{tab:teacher_checkpoints} lists the exact model IDs and native image resolutions we use.

\begin{table}[h!]
\centering
\small
\caption{\textbf{Teacher encoders used for alignment.} We adopt each teacher’s native evaluation resolution and keep their weights frozen during alignment.}

\label{tab:teacher_checkpoints}
\setlength{\tabcolsep}{6pt}
\begin{tabular}{@{}l l c l@{}}
\toprule
\textbf{Family} & \textbf{Architecture} & \textbf{Eval res.} & \textbf{Identifier} \\
\midrule

\multicolumn{4}{@{}l}{\textit{Fully Supervised Models} \cite{hf_vit_base_patch16_224}} \\
Supervised~\cite{dosovitskiy2021an} & ViT-B/16 & 224$\times$224 & \texttt{google/vit-base-patch16-224} \\
\midrule

\multicolumn{4}{@{}l}{\textit{Vision--Language Models} \cite{hf_siglip2_base,hf_siglip2_large,hf_siglip2_so400m}} \\
SigLIP2~\cite{tschannen2025siglip} & ViT-B/16   & 224$\times$224 & \texttt{google/siglip2-base-patch16-224} \\
SigLIP2~\cite{tschannen2025siglip} & ViT-L/16   & 256$\times$256 & \texttt{google/siglip2-large-patch16-256} \\
SigLIP2~\cite{tschannen2025siglip}& ViT-so400m/16  & 256$\times$256 & \texttt{google/siglip2-so400m-patch16-256} \\
SigLIP2~\cite{tschannen2025siglip}& ViT-so400m/14@384  & 384$\times$384 & \texttt{google/siglip2-so400m-patch14-384} \\
\midrule

\multicolumn{4}{@{}l}{\textit{Self-Supervised Models} \cite{hf_dinov3_vits,hf_dinov3_vitb,hf_dinov3_vitl}} \\
DINOv3~\cite{simeoni2025dinov3} & ViT-S/16 & 224$\times$224 & \texttt{facebook/dinov3-vits16-pretrain-lvd1689m} \\
DINOv3~\cite{simeoni2025dinov3} & ViT-B/16 & 224$\times$224 & \texttt{facebook/dinov3-vitb16-pretrain-lvd1689m} \\
DINOv3~\cite{simeoni2025dinov3}& ViT-L/16 & 224$\times$224 & \texttt{facebook/dinov3-vitl16-pretrain-lvd1689m} \\
\bottomrule
\end{tabular}

\end{table}

\myparagraph{Input resolutions.}
Alignment and evaluation follow the teacher’s native resolution (Table~\ref{tab:teacher_checkpoints}): 224$\times$224 for Supervised ViT-B/16 and all DINOv3 models; 224$\times$224 for SigLIP2-B/16; 256$\times$256 for SigLIP2-L/16 and SigLIP2-so400m/16, and 384$\times$384 for SigLIP2-so400m/14@384.
For the conv-stem \bcos counterpart of SigLIP2-so400m/14@384, replacing the patch embedding by a stride-16 convolution changes the effective patch size from 14 to 16; we therefore evaluate this aligned model at 432$\times$432 and denote it as B-ViT-so400m/16@432 in the tables.

\myparagraph{Evaluation datasets.}
Linear Probing (LP) and $k$-NN share the same 10 datasets: \textsc{IN1k}~\cite{imagenet}, \textsc{Caltech101}~\cite{caltech101}, \textsc{Flowers102}~\cite{nilsback2008automated}, \textsc{Food101}~\cite{bossard2014food}, \textsc{FGVC-Aircraft}~\cite{maji2013fine}, \textsc{DTD}~\cite{cimpoi2014describing}, \textsc{Stanford Cars}~\cite{krause20133d}, \textsc{SUN397}~\cite{xiao2010sun}, \textsc{CIFAR-10}~\cite{krizhevsky2009learning}, \textsc{CIFAR-100}~\cite{krizhevsky2009learning}. Dense linear probing for depth uses \textsc{NYUv2}~\cite{silberman2012indoor}.

\myparagraph{Dense linear probing for depth.} We train a \emph{linear} depth head on frozen features, and optimize with an L1 objective on inverse depth plus a scale-invariant gradient prior, following the Probe3D~\cite{el2024probing} protocol. Inputs are resized to the teacher’s native resolution and center-cropped; no test-time augmentation is used. We evaluate on the standard \textsc{NYUv2}~\cite{silberman2012indoor} split and report both \textbf{relative} and \textbf{absolute} metrics: $\delta_{1}\uparrow$ (fraction of pixels satisfying $\max(\hat d/d, d/\hat d)<1.25$, where $\hat d$ is predicted depth and $d$ is ground-truth depth; higher is better) and RMSE$\downarrow$ (lower is better).

\myparagraph{Surface normals estimation.}
We also evaluate surface normal estimation on \textsc{NYUv2}~\cite{silberman2012indoor,ladicky2014discriminatively} following the Probe3D~\cite{el2024probing} protocol by training a linear head on frozen features, and report $\delta_1$, $\delta_2$, $\delta_3$, and RMSE.

\myparagraph{Multiview correspondence.}
We evaluate multiview correspondence with Probe3D~\cite{el2024probing} on two different datasets: NAVI~\cite{jampani2023navi} and ScanNet~\cite{dai2017scannet}, reporting recall at standard error thresholds over view pairs from the same 3D scene.

\myparagraph{Zero-shot (SigLIP2~\cite{tschannen2025siglip}).}
We replace only the \emph{image} encoder with its ALOE-aligned \bcos counterpart; the SigLIP2 \emph{text} encoder, prompt templates, temperature, and normalization follow the originals~\cite{radford2021learning,tschannen2025siglip}. We report ImageNet~\cite{imagenet} top-1 accuracy with the OpenCLIP 80-prompt template (\cite{ilharco_gabriel_2021_5143773,cherti2023reproducible}), averaging text embeddings over the prompts of each class. Zero-shot explanations use the single prompt ``this is a photo of \{class name\}''.

\myparagraph{Multimodal large language model token grounding.}
For the multimodal large language model (MLLM) examples, we pair our ALOE-aligned SigLIP2 \bcos encoder with a LLaVA-MORE~\cite{cocchi2025llava} Gemma-2-9B~\cite{gemmateam2024gemma2improvingopen} language backbone. Relevance is propagated through the language model with AttnLRP~\cite{achtibat2024attnlrp}, after which we extract model-inherent visual explanations from the \bcos encoder to obtain token-level grounding maps.

\myparagraph{Attribution methods visualized.}
For the inherently interpretable \bcos models~\cite{boehle2024bcos}, explanations are model-inherent $\mathbf{W}(\mathbf{x})\mathbf{x}$~\cite{boehle2022bcos}. For conventional teachers, we visualize AttnLRP~\cite{achtibat2024attnlrp}, LeGrad~\cite{bousselham2024legrad} and CheferCAM~\cite{chefer2021transformer} (using the authors' original implementation), and Integrated Gradients~\cite{sundararajan2017axiomatic} as well as LIME~\cite{ribeiro2016should} (from the \texttt{captum} library~\cite{kokhlikyan2020captum}). Where applicable, we use authors’ recommended defaults.

\myparagraph{Interpretability metrics.}
We evaluate model attributions with (i) the \textit{Grid Pointing Game (GridPG)}~\cite{boehle2021convolutional,zhang2018top} for localization and (ii) \textit{Pixel Deletion}~\cite{samek2016evaluating} for faithfulness, following standard protocol from prior work.

\noindent\textit{GridPG.}
We build $N{\times}N$ grids (we use $2{\times}2$) from images of \emph{distinct} classes that are individually and confidently correctly classified. For each class $i$, we measure the fraction of \emph{positive} attribution mass inside its corresponding grid cell. Let $A(p)$ be the attribution at pixel $p$ and $A^+(p)=\max(A(p),0)$ its positive part; the localization score for cell $i$ is
\[
L_i \;=\; \frac{\sum_{p \in \text{cell}_i} A^+(p)}{\sum_{j=1}^{N^2} \sum_{p \in \text{cell}_j} A^+(p)}\,,
\]
and the GridPG score is the average of $L_i$ over several grids (grids with zero total positive mass are discarded).

\noindent\textit{Pixel Deletion.}
We rank $16\times16$ pixel blocks by attribution score from most to least important and iteratively set the pixels in the highest-ranked blocks to zero, plotting the target-class probability versus the fraction of blocks removed; \emph{larger} drops (steeper curves) indicate attributions that are more consistent with model decisions.

\section{Earlier Checkpoints and Dagger Notation}
\label{supp:sec:prevreg}

This arXiv version uses updated ALOE checkpoints trained with the updated B-cos implementation and without the temporary weight-norm stabilization term for all unmarked results (without \prevreg). This includes the main classification, zero-shot, dense-prediction, data-scaling, localization, and faithfulness results. Compared with the CVPR-version checkpoints~\cite{Maser_2026_CVPR}, the updated checkpoints slightly improve recognition performance while leaving explanation quality essentially unchanged.

Results marked with \prevreg use earlier CVPR-version ALOE checkpoints~\cite{Maser_2026_CVPR}. In this version, the remaining earlier-checkpoint material is the alignment-objective and feature-depth ablations (\cref{tab:loss_ablation_in1k_lp,tab:where_loss_cols}), the token-level LVLM visual-grounding examples (main \cref{fig:llava_grounding} and supplementary \cref{fig:vlm_teaser_supplement_1,fig:vlm_teaser_supplement_2,fig:vlm_teaser_supplement_3}), and the human user-study material (\cref{fig:human_study_example,tab:human_eval_styled}). 

The earlier checkpoints were trained with a temporary weight-norm stabilization term,
\begin{equation}
    \mathcal{L}_{\mathrm{reg}} = \alpha \sum_{\ell \in \mathcal{P}} \left( \| \mathbf{W}^{(T)}_\ell \|_F - \| \mathbf{W}^{(S)}_\ell \|_F \right)^2,
\end{equation}
to prevent potentially unstable weight-norm growth in longer training runs. This potential instability stems from the scaling of the unnormalized B-cos transform in the public B-cosification code\footnote{\url{https://github.com/shrebox/B-cosification}}. For an activation $\mathbf{x}$ and weight row $\mathbf{w}_j$, the legacy scale used
\begin{equation}
    \left(\frac{|\mathbf{w}_j^\top \mathbf{x}|}{\|\mathbf{x}\|_2}\right)^{B-1}
\end{equation}
instead of the unnormalized-weight form
\begin{equation}
    \left(\frac{|\mathbf{w}_j^\top \mathbf{x}|}{\|\mathbf{x}\|_2\|\mathbf{w}_j\|_2}\right)^{B-1}
    = |\cos(\mathbf{x},\mathbf{w}_j)|^{B-1}.
\end{equation}
Thus the legacy path scales the dynamic B-cos factor by an additional $\|\mathbf{w}_j\|_2^{B-1}$ (one extra weight-norm factor for $B=2$). The layer remains dynamic-linear, so explanations stay exact, but outputs and gradients become coupled to the weight norms, which can destabilize training when these grow. Our released implementation keeps both paths for reproducibility, and all updated arXiv checkpoints use the unnormalized-weight form. The stabilization term was used for stability, not as a performance-improving regularizer.

\myparagraph{B-cosification baselines.} All B-cosification~\cite{arya2024b} baselines are trained with the original public implementation, \ie, with the legacy scale above, matching the published method. They are therefore not marked with \prevreg.

\clearpage

\section{Additional Quantitative Results}
\label{supp:sec:quantitative}

\begin{table*}[ht!]
\centering
\scriptsize
\setlength{\tabcolsep}{1.0pt}
\caption{\textbf{Linear-probe accuracy on frozen features.} \ours B-cos ViTs substantially outperform B-cosification while remaining competitive with the original foundation models on ImageNet-1k and on the 10-dataset average. All models use the same protocol and resolution. Teachers are shown in gray; best per block in \textbf{bold} (for \bcos models where a vanilla B-cosification baseline exists). \cmark: denotes inherently interpretable models (vs. not \xmark).}
\label{tab:lp_supp}

\begin{tabularx}{\textwidth}{@{} l c c Y Y Y Y Y Y Y Y Y Y Y @{}}
\toprule
\textbf{Feature} & 
\textbf{Arch} & 
\rotatebox{\rotangle}{\textbf{Inter.}} & 
\rotatebox{\rotangle}{\textbf{IN1k}} & 
\rotatebox{\rotangle}{\textbf{Cal101}} & 
\rotatebox{\rotangle}{\textbf{Flowers}} & 
\rotatebox{\rotangle}{\textbf{Food}} & 
\rotatebox{\rotangle}{\textbf{Aircr}} & 
\rotatebox{\rotangle}{\textbf{DTD}} & 
\rotatebox{\rotangle}{\textbf{Cars}} & 
\rotatebox{\rotangle}{\textbf{SUN}} & 
\rotatebox{\rotangle}{\textbf{C10}} & 
\rotatebox{\rotangle}{\textbf{C100}} & 
\rotatebox{\rotangle}{\textbf{Avg.}} \\
\midrule

\multicolumn{14}{c}{\textit{Fully Supervised Pre-Training}} \\
\midrule
\textcolor{mygray}{Sup.~\cite{dosovitskiy2021an}} & \textcolor{mygray}{ViT-B/16} & \xmark & \textcolor{mygray}{81.19} & \textcolor{mygray}{97.43} & \textcolor{mygray}{99.33} & \textcolor{mygray}{86.22} & \textcolor{mygray}{41.31} & \textcolor{mygray}{74.44} & \textcolor{mygray}{57.39} & \textcolor{mygray}{74.19} & \textcolor{mygray}{97.07} & \textcolor{mygray}{86.56} & \textcolor{mygray}{79.51} \\

B-cosif.~\cite{arya2024b} & B-ViT-B/16 & \cmark & 77.65 & 96.35 & 95.83 & 78.18 & 37.07 & 71.15 & 46.06 & 66.83 & 95.00 & 81.43 & 74.56 \\

\rowcolor{hirow}
\ours & B-ViT-B/16 & \cmark & \textbf{81.30} & \textbf{97.32} & \textbf{99.44} & \textbf{86.73} & \textbf{43.96} & \textbf{75.55} & \textbf{61.59} & \textbf{74.99} & \textbf{97.53} & \textbf{87.67} & \textbf{80.61} \\

& & & \textcolor{teal}{+3.65} & \textcolor{teal}{+0.97} & \textcolor{teal}{+3.61} & \textcolor{teal}{+8.55} & \textcolor{teal}{+6.89} & \textcolor{teal}{+4.40} & \textcolor{teal}{+15.53} & \textcolor{teal}{+8.16} & \textcolor{teal}{+2.53} & \textcolor{teal}{+6.24} & \textcolor{teal}{+6.05} \\

\midrule
\multicolumn{14}{c}{\textit{Vision Language Pre-training}} \\
\midrule
\textcolor{mygray}{SigLIP2~\cite{tschannen2025siglip}} & \textcolor{mygray}{ViT-B/16} & \xmark & \textcolor{mygray}{84.21} & \textcolor{mygray}{98.88} & \textcolor{mygray}{99.21} & \textcolor{mygray}{94.44} & \textcolor{mygray}{75.48} & \textcolor{mygray}{85.74} & \textcolor{mygray}{95.42} & \textcolor{mygray}{81.70} & \textcolor{mygray}{96.95} & \textcolor{mygray}{84.91} & \textcolor{mygray}{89.69} \\

B-cosif.~\cite{arya2024b} & B-ViT-B/16 & \cmark & 75.05 & 97.52 & 94.14 & 83.74 & 45.04 & 79.18 & 73.70 & 75.68 & 94.44 & 81.11 & 79.96 \\

\rowcolor{hirow}
\ours & B-ViT-B/16 & \cmark & \textbf{83.87} & \textbf{98.32} & \textbf{98.84} & \textbf{94.20} & \textbf{72.47} & \textbf{83.03} & \textbf{94.31} & \textbf{81.52} & \textbf{96.71} & \textbf{84.53} & \textbf{88.78} \\

& & & \textcolor{teal}{+8.82} & \textcolor{teal}{+0.80} & \textcolor{teal}{+4.70} & \textcolor{teal}{+10.46} & \textcolor{teal}{+27.43} & \textcolor{teal}{+3.85} & \textcolor{teal}{+20.61} & \textcolor{teal}{+5.84} & \textcolor{teal}{+2.27} & \textcolor{teal}{+3.42} & \textcolor{teal}{+8.82} \\

\hline
\multicolumn{14}{l}{\textit{Larger Architectures}} \\
\textcolor{mygray}{SigLIP2~\cite{tschannen2025siglip}} & \textcolor{mygray}{ViT-L/16} & \xmark & \textcolor{mygray}{87.15} & \textcolor{mygray}{98.99} & \textcolor{mygray}{99.44} & \textcolor{mygray}{96.44} & \textcolor{mygray}{84.25} & \textcolor{mygray}{87.61} & \textcolor{mygray}{96.33} & \textcolor{mygray}{84.11} & \textcolor{mygray}{97.83} & \textcolor{mygray}{87.77} & \textcolor{mygray}{91.99} \\
\textcolor{mygray}{SigLIP2~\cite{tschannen2025siglip}} & \textcolor{mygray}{ViT-so400m/16} & \xmark & \textcolor{mygray}{87.98} & \textcolor{mygray}{99.10} & \textcolor{mygray}{99.77} & \textcolor{mygray}{97.01} & \textcolor{mygray}{84.04} & \textcolor{mygray}{88.56} & \textcolor{mygray}{96.61} & \textcolor{mygray}{84.67} & \textcolor{mygray}{98.58} & \textcolor{mygray}{89.76} & \textcolor{mygray}{92.61} \\
\textcolor{mygray}{SigLIP2~\cite{tschannen2025siglip}} & \textcolor{mygray}{ViT-so400m/14@384} & \xmark & \textcolor{mygray}{88.65} & \textcolor{mygray}{99.21} & \textcolor{mygray}{99.77} & \textcolor{mygray}{97.45} & \textcolor{mygray}{85.09} & \textcolor{mygray}{88.67} & \textcolor{mygray}{96.81} & \textcolor{mygray}{85.23} & \textcolor{mygray}{98.52} & \textcolor{mygray}{89.60} & \textcolor{mygray}{92.90} \\

\rowcolor{hirow}
\ours & B-ViT-L/16 & \cmark & 87.08 & 98.77 & 99.55 & 96.38 & 81.85 & 87.05 & 96.20 & 84.23 & 98.04 & 87.79 & 91.69 \\
\rowcolor{hirow}
\ours & B-ViT-so400m/16 & \cmark & 87.83 & 99.10 & 99.77 & 96.87 & 82.48 & 88.33 & 96.38 & 84.51 & 98.56 & 89.91 & 92.37 \\
\rowcolor{hirow}
\ours & B-ViT-so400m/16@432 & \cmark & 88.48 & 99.21 & 99.77 & 97.37 & 83.17 & 89.00 & 96.74 & 85.07 & 98.43 & 89.74 & 92.70 \\

\midrule
\multicolumn{14}{c}{\textit{Self-Supervised Pre-training}} \\
\midrule
\multicolumn{14}{l}{\textit{Smaller Architectures}} \\
\textcolor{mygray}{DINOv3~\cite{simeoni2025dinov3}} & \textcolor{mygray}{ViT-S/16} & \xmark & \textcolor{mygray}{78.62} & \textcolor{mygray}{97.99} & \textcolor{mygray}{99.77} & \textcolor{mygray}{89.59} & \textcolor{mygray}{72.89} & \textcolor{mygray}{80.85} & \textcolor{mygray}{91.59} & \textcolor{mygray}{74.41} & \textcolor{mygray}{96.25} & \textcolor{mygray}{85.04} & \textcolor{mygray}{86.70} \\
\rowcolor{hirow}
\ours & B-ViT-S/16 & \cmark & 77.91 & 98.10 & 99.77 & 88.82 & 71.75 & 80.02 & 91.30 & 73.88 & 95.50 & 84.26 & 86.13 \\

\hline
\multicolumn{14}{l}{\textit{Base Architecture}} \\
\textcolor{mygray}{DINOv3~\cite{simeoni2025dinov3}} & \textcolor{mygray}{ViT-B/16} & \xmark & \textcolor{mygray}{84.36} & \textcolor{mygray}{98.88} & \textcolor{mygray}{99.55} & \textcolor{mygray}{94.12} & \textcolor{mygray}{80.40} & \textcolor{mygray}{83.98} & \textcolor{mygray}{94.51} & \textcolor{mygray}{78.74} & \textcolor{mygray}{98.22} & \textcolor{mygray}{89.24} & \textcolor{mygray}{90.20} \\

B-cosif.~\cite{arya2024b} & B-ViT-B/16 & \cmark & 73.64 & 95.18 & 82.68 & 68.24 & 41.67 & 68.02 & 50.90 & 58.25 & 92.04 & 76.44 & 70.71 \\

\rowcolor{hirow}
\ours & B-ViT-B/16 & \cmark & \textbf{84.14} & \textbf{98.77} & \textbf{99.55} & \textbf{93.86} & \textbf{80.19} & \textbf{83.42} & \textbf{94.41} & \textbf{78.23} & \textbf{98.01} & \textbf{89.35} & \textbf{89.99} \\

& & & \textcolor{teal}{+10.50} & \textcolor{teal}{+3.59} & \textcolor{teal}{+16.87} & \textcolor{teal}{+25.62} & \textcolor{teal}{+38.52} & \textcolor{teal}{+15.40} & \textcolor{teal}{+43.51} & \textcolor{teal}{+19.98} & \textcolor{teal}{+5.97} & \textcolor{teal}{+12.91} & \textcolor{teal}{+19.28} \\

\hline
\multicolumn{14}{l}{\textit{Larger Architectures}} \\
\textcolor{mygray}{DINOv3~\cite{simeoni2025dinov3}} & \textcolor{mygray}{ViT-L/16} & \xmark & \textcolor{mygray}{87.08} & \textcolor{mygray}{98.77} & \textcolor{mygray}{99.66} & \textcolor{mygray}{95.91} & \textcolor{mygray}{80.22} & \textcolor{mygray}{87.33} & \textcolor{mygray}{94.74} & \textcolor{mygray}{80.82} & \textcolor{mygray}{99.14} & \textcolor{mygray}{93.27} & \textcolor{mygray}{91.69} \\
\rowcolor{hirow}
\ours & B-ViT-L/16 & \cmark & 86.82 & 98.77 & 99.66 & 95.87 & 79.14 & 86.32 & 94.82 & 80.32 & 99.16 & 92.99 & 91.39 \\

\bottomrule
\end{tabularx}
\end{table*}

\clearpage
In this section, we expand the quantitative evaluation along two axes: (1) \emph{downstream performance}, and (2) \emph{interpretability/faithfulness}. In~\cref{sec:supp_downstream}, for downstream performance, we report $k$-NN on frozen features (\cref{tab:knn_all}), zero-shot transfer for SigLIP2 (\cref{tab:zeroshot_siglip2}), dense prediction tasks (\cref{tab:dense_depth_vitb16,tab:surface_normals}, \cref{fig:multiview_correspondence}), and data-efficiency analyses (\cref{fig:aloe_dino_data_scaling}). In~\cref{sec:supp_interp}, for interpretability and faithfulness, we quantify localization with GridPG (\cref{tab:gridpg_localization}) and evaluate stability via pixel-deletion tests (\cref{fig:perturb_stability}). Additionally, we also report results on a human preference study. Unless noted otherwise, protocols match the main paper (see \cref{sec:aloe_eval}) and \cref{supp:sec:impl}.

\begin{table*}[ht]
\centering
\scriptsize
\setlength{\tabcolsep}{1.0pt}
\caption{\textbf{$k$-NN accuracy on frozen features.} For the $k$-NN ($k=20$) evaluation setting, \ours B-cos ViTs again significantly outperform B-cosification~\cite{arya2024b} while remaining competitive with the original foundation models on ImageNet-1k and on the 10-dataset average. All models use the same protocol and resolution (see~\cref{sec:aloe_eval}). Teachers are shown in gray; best per block in \textbf{bold} (for \bcos models). \cmark: denotes inherently interpretable models (vs. not \xmark).}
\label{tab:knn_all}

\begin{tabularx}{\textwidth}{@{} l c c Y Y Y Y Y Y Y Y Y Y Y @{}}
\toprule
\textbf{Feature} & 
\textbf{Arch} & 
\rotatebox{\rotangle}{\textbf{Inter.}} & 
\rotatebox{\rotangle}{\textbf{IN1k}} & 
\rotatebox{\rotangle}{\textbf{Cal101}} & 
\rotatebox{\rotangle}{\textbf{Flowers}} & 
\rotatebox{\rotangle}{\textbf{Food}} & 
\rotatebox{\rotangle}{\textbf{Aircr}} & 
\rotatebox{\rotangle}{\textbf{DTD}} & 
\rotatebox{\rotangle}{\textbf{Cars}} & 
\rotatebox{\rotangle}{\textbf{SUN}} & 
\rotatebox{\rotangle}{\textbf{C10}} & 
\rotatebox{\rotangle}{\textbf{C100}} & 
\rotatebox{\rotangle}{\textbf{Avg.}} \\
\midrule

\multicolumn{14}{c}{\textit{Fully Supervised Pre-Training}} \\
\midrule
\textcolor{mygray}{Sup.~\cite{dosovitskiy2021an}} & \textcolor{mygray}{ViT-B/16} & \xmark & \textcolor{mygray}{80.70} & \textcolor{mygray}{92.07} & \textcolor{mygray}{87.61} & \textcolor{mygray}{78.98} & \textcolor{mygray}{22.02} & \textcolor{mygray}{62.89} & \textcolor{mygray}{29.96} & \textcolor{mygray}{68.62} & \textcolor{mygray}{96.38} & \textcolor{mygray}{82.42} & \textcolor{mygray}{70.17} \\

B-cosif.~\cite{arya2024b} & B-ViT-B/16 & \cmark & 77.05 & 89.58 & 75.13 & 65.00 & 19.14 & 59.20 & 23.45 & 56.74 & 93.14 & 74.57 & 63.30 \\

\rowcolor{hirow}
\ours (ours) & B-ViT-B/16 & \cmark & \textbf{80.76} & \textbf{92.29} & \textbf{87.61} & \textbf{79.73} & \textbf{23.04} & \textbf{63.50} & \textbf{31.38} & \textbf{68.98} & \textbf{96.65} & \textbf{83.20} & \textbf{70.71} \\

& & & \textcolor{teal}{+3.71} & \textcolor{teal}{+2.71} & \textcolor{teal}{+12.48} & \textcolor{teal}{+14.73} & \textcolor{teal}{+3.90} & \textcolor{teal}{+4.30} & \textcolor{teal}{+7.93} & \textcolor{teal}{+12.24} & \textcolor{teal}{+3.51} & \textcolor{teal}{+8.63} & \textcolor{teal}{+7.41} \\

\midrule
\multicolumn{14}{c}{\textit{Vision Language Pre-training}} \\
\midrule
\textcolor{mygray}{SigLIP2~\cite{tschannen2025siglip}} & \textcolor{mygray}{ViT-B/16} & \xmark & \textcolor{mygray}{80.39} & \textcolor{mygray}{97.54} & \textcolor{mygray}{89.73} & \textcolor{mygray}{93.21} & \textcolor{mygray}{65.17} & \textcolor{mygray}{76.61} & \textcolor{mygray}{92.41} & \textcolor{mygray}{75.72} & \textcolor{mygray}{95.54} & \textcolor{mygray}{79.84} & \textcolor{mygray}{84.62} \\

B-cosif.~\cite{arya2024b} & B-ViT-B/16 & \cmark & 68.42 & 94.14 & 74.87 & 76.16 & 25.45 & 72.43 & 49.57 & 68.46 & 91.74 & 72.63 & 69.39 \\

\rowcolor{hirow}
\ours (ours) & B-ViT-B/16 & \cmark & \textbf{79.97} & \textbf{97.09} & \textbf{89.06} & \textbf{92.62} & \textbf{62.65} & \textbf{76.06} & \textbf{90.46} & \textbf{75.49} & \textbf{95.19} & \textbf{79.47} & \textbf{83.81} \\

& & & \textcolor{teal}{+11.55} & \textcolor{teal}{+2.95} & \textcolor{teal}{+14.19} & \textcolor{teal}{+16.46} & \textcolor{teal}{+37.20} & \textcolor{teal}{+3.63} & \textcolor{teal}{+40.89} & \textcolor{teal}{+7.03} & \textcolor{teal}{+3.45} & \textcolor{teal}{+6.84} & \textcolor{teal}{+14.42} \\

\hline
\multicolumn{14}{l}{\textit{Larger Architectures}} \\
\textcolor{mygray}{SigLIP2~\cite{tschannen2025siglip}} & \textcolor{mygray}{ViT-L/16} & \xmark & \textcolor{mygray}{83.76} & \textcolor{mygray}{97.87} & \textcolor{mygray}{94.97} & \textcolor{mygray}{95.55} & \textcolor{mygray}{73.79} & \textcolor{mygray}{78.06} & \textcolor{mygray}{93.47} & \textcolor{mygray}{77.17} & \textcolor{mygray}{96.61} & \textcolor{mygray}{82.62} & \textcolor{mygray}{87.39} \\
\textcolor{mygray}{SigLIP2~\cite{tschannen2025siglip}} & \textcolor{mygray}{ViT-so400m/16} & \xmark & \textcolor{mygray}{84.51} & \textcolor{mygray}{97.87} & \textcolor{mygray}{91.07} & \textcolor{mygray}{96.12} & \textcolor{mygray}{74.42} & \textcolor{mygray}{78.29} & \textcolor{mygray}{93.56} & \textcolor{mygray}{77.63} & \textcolor{mygray}{97.74} & \textcolor{mygray}{84.85} & \textcolor{mygray}{87.61} \\
\textcolor{mygray}{SigLIP2~\cite{tschannen2025siglip}} & \textcolor{mygray}{ViT-so400m/14@384} & \xmark & \textcolor{mygray}{85.08} & \textcolor{mygray}{98.32} & \textcolor{mygray}{90.29} & \textcolor{mygray}{96.61} & \textcolor{mygray}{72.98} & \textcolor{mygray}{78.68} & \textcolor{mygray}{94.01} & \textcolor{mygray}{77.42} & \textcolor{mygray}{97.62} & \textcolor{mygray}{84.74} & \textcolor{mygray}{87.58} \\

\rowcolor{hirow}
\ours (ours) & B-ViT-L/16 & \cmark & {83.89} & {97.21} & {89.73} & {95.53} & {71.54} & {76.56} & {92.79} & {77.74} & {96.75} & {83.09} & {86.48} \\
\rowcolor{hirow}
\ours (ours) & B-ViT-so400m/16 & \cmark & {84.56} & {97.43} & {90.73} & {96.07} & {71.30} & {79.12} & {93.12} & {77.63} & {97.85} & {85.50} & {87.33} \\
\rowcolor{hirow}
\ours (ours) & B-ViT-so400m/16@432 & \cmark & {85.13} & {97.21} & {91.51} & {96.55} & {72.44} & {77.62} & {93.34} & {77.54} & {97.60} & {84.90} & {87.38} \\

\midrule
\multicolumn{14}{c}{\textit{Self-Supervised Pre-training}} \\
\midrule
\multicolumn{14}{l}{\textit{Smaller Architectures}} \\
\textcolor{mygray}{DINOv3~\cite{simeoni2025dinov3}} & \textcolor{mygray}{ViT-S/16} & \xmark & \textcolor{mygray}{76.92} & \textcolor{mygray}{94.30} & \textcolor{mygray}{93.52} & \textcolor{mygray}{85.89} & \textcolor{mygray}{51.77} & \textcolor{mygray}{75.50} & \textcolor{mygray}{83.74} & \textcolor{mygray}{67.97} & \textcolor{mygray}{95.73} & \textcolor{mygray}{81.55} & \textcolor{mygray}{80.69} \\
\rowcolor{hirow}
\ours (ours) & B-ViT-S/16 & \cmark & {75.83} & {93.63} & {92.52} & {84.59} & {50.24} & {74.66} & {82.68} & {66.94} & {95.25} & {80.10} & {79.64} \\

\hline
\multicolumn{14}{l}{\textit{Base Architecture}} \\
\textcolor{mygray}{DINOv3~\cite{simeoni2025dinov3}} & \textcolor{mygray}{ViT-B/16} & \xmark & \textcolor{mygray}{82.28} & \textcolor{mygray}{94.97} & \textcolor{mygray}{87.72} & \textcolor{mygray}{91.37} & \textcolor{mygray}{58.59} & \textcolor{mygray}{77.84} & \textcolor{mygray}{88.94} & \textcolor{mygray}{72.44} & \textcolor{mygray}{97.32} & \textcolor{mygray}{85.67} & \textcolor{mygray}{83.71} \\

B-cosif.~\cite{arya2024b} & B-ViT-B/16 & \cmark & 71.03 & 83.33 & 41.27 & 44.97 & 16.25 & 44.64 & 22.90 & 34.86 & 87.32 & 61.18 & 50.78 \\

\rowcolor{hirow}
\ours (ours) & B-ViT-B/16 & \cmark & \textbf{81.56} & \textbf{94.30} & \textbf{88.05} & \textbf{90.49} & \textbf{58.11} & \textbf{77.56} & \textbf{88.24} & \textbf{72.15} & \textbf{97.61} & \textbf{85.75} & \textbf{83.38} \\

& & & \textcolor{teal}{+10.53} & \textcolor{teal}{+10.97} & \textcolor{teal}{+46.78} & \textcolor{teal}{+45.52} & \textcolor{teal}{+41.86} & \textcolor{teal}{+32.92} & \textcolor{teal}{+65.34} & \textcolor{teal}{+37.29} & \textcolor{teal}{+10.29} & \textcolor{teal}{+24.57} & \textcolor{teal}{+32.60} \\

\hline
\multicolumn{14}{l}{\textit{Larger Architectures}} \\
\textcolor{mygray}{DINOv3~\cite{simeoni2025dinov3}} & \textcolor{mygray}{ViT-L/16} & \xmark & \textcolor{mygray}{84.77} & \textcolor{mygray}{94.19} & \textcolor{mygray}{83.48} & \textcolor{mygray}{93.79} & \textcolor{mygray}{57.42} & \textcolor{mygray}{77.06} & \textcolor{mygray}{90.61} & \textcolor{mygray}{74.38} & \textcolor{mygray}{98.47} & \textcolor{mygray}{90.01} & \textcolor{mygray}{84.42} \\
\rowcolor{hirow}
\ours (ours) & B-ViT-L/16 & \cmark & {83.71} & {93.52} & {91.18} & {93.26} & {56.73} & {76.45} & {90.23} & {72.56} & {99.03} & {90.46} & {84.71} \\

\bottomrule
\end{tabularx}
\end{table*}

\subsection{Downstream performance}
\label{sec:supp_downstream}

\myparagraph{LP on frozen features.}
Across ten datasets, \ours B\mbox{-}ViTs substantially outperform B\mbox{-}cosification while remaining competitive with their teachers (\cref{tab:lp_supp}). Gains are especially pronounced on fine-grained or texture-heavy benchmarks (e.g., \textsc{Cars}, \textsc{Aircr}, \textsc{DTD}, \textsc{Food}).

\myparagraph{$k$-NN on frozen features.}
Similar to the LP performance \cref{tab:lp_supp} our models substantially outperform B\mbox{-}cosification while preserving most of the baseline's capabilities, indicating that alignment preserves discriminative structure in feature space without additional fine-tuning (see \cref{tab:knn_all}).

\begin{table}[t!]
\centering
\small
\caption{\textbf{Zero-shot ImageNet-1k with SigLIP2 prompts.} We replace the SigLIP2 image encoder with the ALOE-aligned \bcos counterpart and evaluate zero-shot classification with the OpenCLIP 80-prompt template~\cite{ilharco_gabriel_2021_5143773,cherti2023reproducible}. Values are ImageNet top-1 accuracy (\%). Teachers are shown in gray; \cmark~denotes inherently interpretable \bcos models (\xmark~are not). For ViT-B/16, \ours substantially outperforms B\mbox{-}cosification and remains competitive with the teacher; similar trends hold for larger models. The \textcolor{teal}{$\Delta$} (\textbf{row 4}) reports \ours{} minus B-cosification.}
\label{tab:zeroshot_siglip2}

\setlength{\tabcolsep}{5pt}
\begin{tabular}{@{}lcc@{}}
\toprule
\textbf{Architecture} & \textbf{Inh. Inter.} & \textbf{Zero-shot Acc.} \\
\midrule
\multicolumn{3}{c}{\textit{Base architecture}} \\
\midrule
\textcolor{mygray}{SigLIP2~\cite{tschannen2025siglip} — ViT-B/16} & \xmark & \textcolor{mygray}{78.07} \\
B\mbox{-}cosif.~\cite{arya2024b} — B\mbox{-}ViT\mbox{-}B/16 & \cmark & 58.66 \\
\rowcolor{hirow}
ALOE (ours) — B\mbox{-}ViT\mbox{-}B/16 & \cmark & \textbf{77.17} \\
\textit{$\Delta$ (\ours vs.\ B\mbox{-}cosif.)} & & \textcolor{teal}{+18.51} \\
\midrule
\multicolumn{3}{c}{\textit{Larger architectures}} \\
\midrule
\textcolor{mygray}{SigLIP2~\cite{tschannen2025siglip} — ViT-L/16} & \xmark & \textcolor{mygray}{82.26} \\
\textcolor{mygray}{SigLIP2~\cite{tschannen2025siglip} — ViT-so400m/16} & \xmark & \textcolor{mygray}{82.56} \\
\textcolor{mygray}{SigLIP2~\cite{tschannen2025siglip} — ViT-so400m/14@384} & \xmark & \textcolor{mygray}{83.54} \\
\rowcolor{hirow}
ALOE (ours) — B\mbox{-}ViT\mbox{-}L/16 & \cmark & \textbf{81.87} \\
\rowcolor{hirow}
ALOE (ours) — B\mbox{-}ViT\mbox{-}so400m/16 & \cmark & \textbf{82.34} \\
\rowcolor{hirow}
ALOE (ours) — B\mbox{-}ViT\mbox{-}so400m/16@432 & \cmark & \textbf{83.25} \\
\bottomrule
\end{tabular}
\end{table}

\myparagraph{Zero-shot (SigLIP2~\cite{tschannen2025siglip}).}
Replacing the SigLIP2 image encoder with its \ours counterpart preserves strong zero-shot classification performance and markedly outperforms B\mbox{-}cosification with the OpenCLIP 80-prompt template (\cite{ilharco_gabriel_2021_5143773,cherti2023reproducible}), while staying close to teacher performance (\cref{tab:zeroshot_siglip2}).

\begin{table}[ht]
\centering
\small
\caption{\textbf{Dense linear probing for monocular depth (ViT-B/16).}
We report relative and absolute depth metrics; higher is better for $\delta_{1}$ and lower is better for RMSE. The \textcolor{teal}{$\Delta$} (\textbf{row 4}) reports \ours{} minus B-cosification.}
\label{tab:dense_depth_vitb16}
\setlength{\tabcolsep}{7pt}
\begin{tabular}{@{}l c rr rr@{}}
\toprule
\multirow{2}{*}{\textbf{Method}} & \multirow{2}{*}{\textbf{Inh.\ Inter.}} & \multicolumn{2}{c}{\textbf{Relative}} & \multicolumn{2}{c}{\textbf{Absolute}} \\
\cmidrule(lr){3-4}\cmidrule(lr){5-6}
& & $\delta_{1}\!\uparrow$ & RMSE$\!\downarrow$ & $\delta_{1}\!\uparrow$ & RMSE$\!\downarrow$ \\
\midrule
\textcolor{mygray}{DINOv3}~\cite{simeoni2025dinov3}  & \xmark & \textcolor{mygray}{0.9542} & \textcolor{mygray}{0.2779} & \textcolor{mygray}{0.7998} & \textcolor{mygray}{0.4748} \\
B\mbox{-}cosif.~\cite{arya2024b}                      & \cmark & 0.8311 & 0.4604 & 0.6503 & 0.6804 \\
\rowcolor{hirow}
\ours (ours)                                          & \cmark & \textbf{0.9316} & \textbf{0.3275} & \textbf{0.7089} & \textbf{0.5845} \\
\addlinespace[2pt]
\textit{$\Delta$ (\ours vs.\ B\mbox{-}cosif.)}       &        & \textcolor{teal}{+0.1005} & \textcolor{teal}{-0.1329} & \textcolor{teal}{+0.0586} & \textcolor{teal}{-0.0959} \\
\bottomrule
\end{tabular}
\end{table}

\begin{table}[ht]
\centering
\small
\caption{\textbf{Surface Normals Estimation (ViT-B/16).} We report angular accuracy metrics ($\delta_n$: fraction of pixels whose angular error is below the $n$-th threshold) and RMSE using Probe3D~\cite{el2024probing}. Higher is better for $\delta$ metrics; lower is better for RMSE. The \textcolor{teal}{$\Delta$} (\textbf{row 4}) reports ALOE minus B-cosification.}
\label{tab:surface_normals}
\setlength{\tabcolsep}{7pt}
\begin{tabular}{@{}l c rrrr@{}}
\toprule
\textbf{Method} & \textbf{Inh.\ Inter.} & $\delta_{1}\!\uparrow$ & $\delta_{2}\!\uparrow$ & $\delta_{3}\!\uparrow$ & RMSE$\!\downarrow$ \\
\midrule
\textcolor{mygray}{DINOv3}~\cite{simeoni2025dinov3} & \xmark & \textcolor{mygray}{0.57} & \textcolor{mygray}{0.77} & \textcolor{mygray}{0.83} & \textcolor{mygray}{23.7} \\
B\mbox{-}cosif. & \cmark & 0.38 & 0.61 & 0.70 & 31.7 \\
\rowcolor{hirow}
\textbf{ALOE (ours)} & \cmark & \textbf{0.54} & \textbf{0.75} & \textbf{0.82} & \textbf{24.9} \\
\addlinespace[2pt]
\textit{$\Delta$ (ALOE vs.\ B\mbox{-}cosif.)} & & \textcolor{teal}{+0.16} & \textcolor{teal}{+0.14} & \textcolor{teal}{+0.12} & \textcolor{teal}{-6.8} \\
\bottomrule
\end{tabular}
\end{table}

\begin{figure}[t]
    \centering
    \begin{subfigure}[b]{0.48\linewidth}
        \centering
        \includegraphics[width=\linewidth]{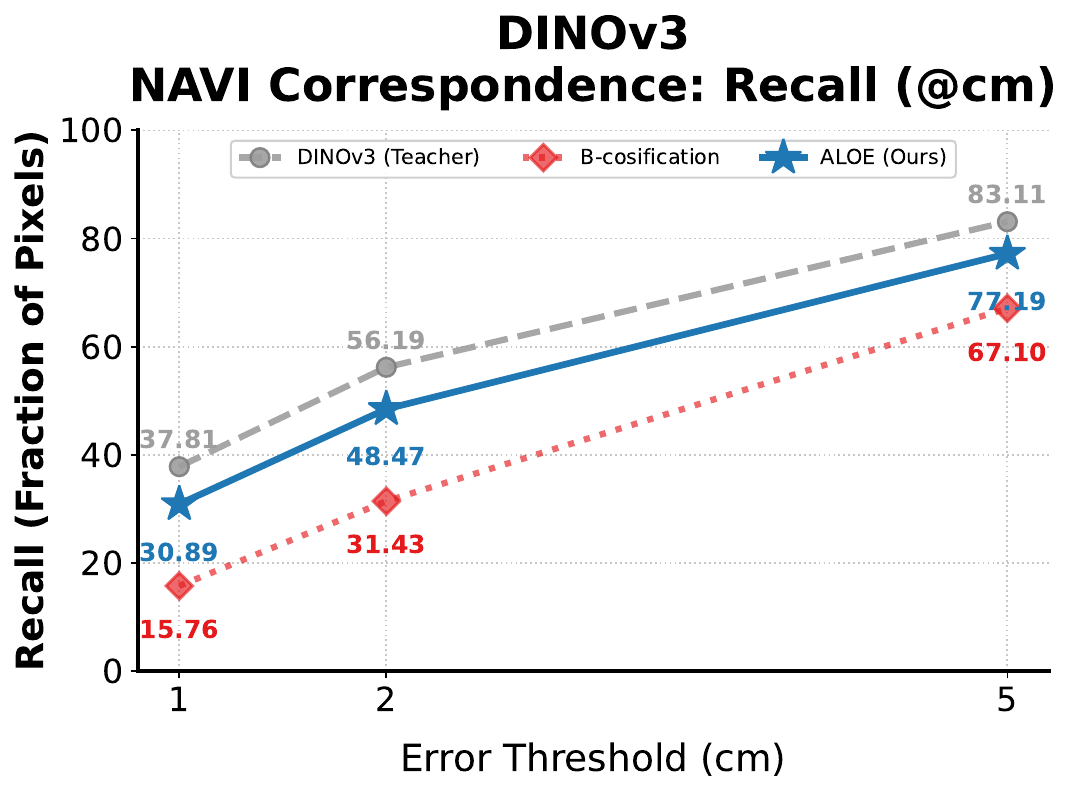}
        \label{fig:multiview_correspondence_navi}
    \end{subfigure}
    \hfill
    \begin{subfigure}[b]{0.48\linewidth}
        \centering
        \includegraphics[width=\linewidth]{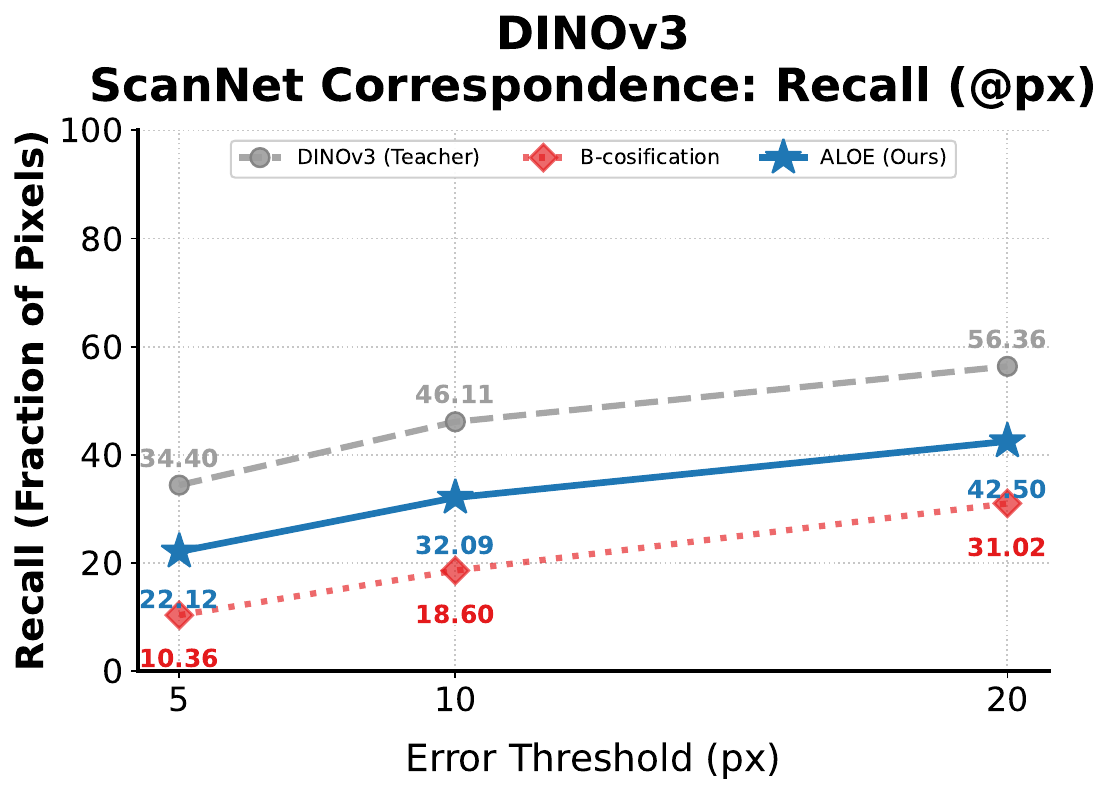}
        \label{fig:multiview_correspondence_scannet}
    \end{subfigure}
    
    \caption{\textbf{Multiview correspondence.} Evaluated on two different datasets (left: NAVI \cite{jampani2023navi}, right: ScanNet \cite{dai2017scannet}) using Probe3D \cite{el2024probing}. \ours increases performance substantially over B-cosification and narrows the gap to the teacher model.}
        \label{fig:multiview_correspondence}
\end{figure}

\myparagraph{Dense linear probing (depth estimation).}
On monocular depth estimation with ViT-B/16, \ours approaches the teacher and surpasses B\mbox{-}cosification on both relative and absolute metrics (\cref{tab:dense_depth_vitb16}) by a large margin. This indicates that aligned \bcos features remain useful for dense prediction, not only global image classification. 

\myparagraph{Surface normal estimation.}
On surface normal estimation, \ours again approaches the teacher and clearly outperforms B\mbox{-}cosification across all angular accuracy metrics and RMSE (\cref{tab:surface_normals}). In particular, \ours improves over B\mbox{-}cosification by \textbf{+0.16}, \textbf{+0.14}, and \textbf{+0.12} on $\delta_1$, $\delta_2$, and $\delta_3$, respectively, while reducing RMSE by \textbf{6.8}. This further indicates that aligned \bcos features preserve spatial structure useful for dense prediction beyond image-level recognition.

\myparagraph{Multiview correspondence.} Multiview correspondence attempts to match two images with different viewing angles from the same 3D scene.
Using the Probe3D framework~\cite{el2024probing}, we also evaluated the multiview correspondence on two different datasets (NAVI \cite{jampani2023navi} and ScanNet \cite{dai2017scannet}). The results are averaged over the different viewing angles and are shown for pre-defined error margins (1, 2, and 5\,cm for NAVI; 5, 10, and 20\,px for ScanNet). \Cref{fig:multiview_correspondence} shows the evaluation results for this experiment.

\myparagraph{Data efficiency.} In addition to SigLIP2 (as demonstrated in \cref{fig:aloe_siglip_data_scaling} of the main paper), we report data-scaling results for DINOv3 that shows a similar trend (\cref{fig:aloe_dino_data_scaling}).

\begin{figure}[h!]
  \centering
  \includegraphics[width=0.8\linewidth]{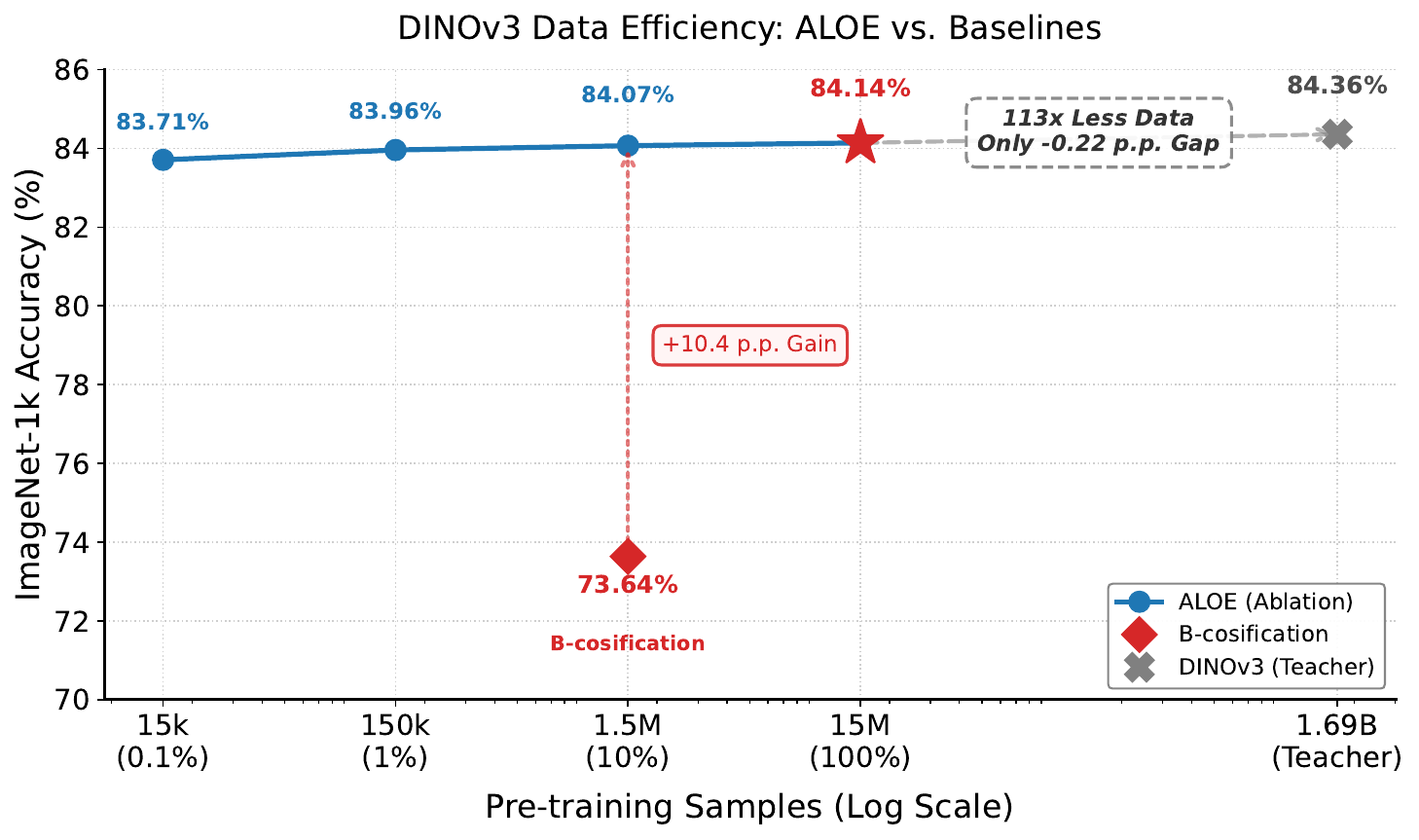}
  \caption{\textbf{Data efficiency of ALOE for DINOv3 on YFCC15M (ViT-B/16).} We vary the amount of unlabeled YFCC15M data used for label-free alignment from 0.1\% (15k images) to 100\% (15M images) and report ImageNet-1k linear-probe accuracy. Across these settings, including the full-data distillation, ALOE remains nearly flat (83.71\%--84.14\%). At 10\% (1.5M images), it already substantially outperforms B-cosification (+10.4 p.p.). The 100\% setting reaches 84.14\%, leaving only a 0.22 p.p. gap to the DINOv3 teacher (84.36\%).
}

  \label{fig:aloe_dino_data_scaling}
\end{figure}

\clearpage
\subsection{Interpretability and faithfulness}
\label{sec:supp_interp}

\begin{table*}[t]
\centering
\footnotesize
\setlength{\tabcolsep}{4pt}
\resizebox{\linewidth}{!}{%
\begin{tabular}{@{}l rr rr rrrrr r@{}}
\toprule
\multirow{2}{*}{\textbf{Model}} & \multicolumn{2}{c}{\textbf{Lin. Probe} \scriptsize{$\uparrow$}} & \multicolumn{2}{c}{\textbf{$k$-NN} \scriptsize{$\uparrow$}} & \multicolumn{5}{c}{\textbf{Grid-PG} \scriptsize{$\uparrow$}} & \multirow{2}{*}{$\boldsymbol{\Delta_{\text{GridPG}}}$} \\
\cmidrule(lr){2-3}\cmidrule(lr){4-5}\cmidrule(lr){6-10}
& Teacher \scriptsize{(\xmark)} & \ours \scriptsize{(\cmark)} & Teacher \scriptsize{(\xmark)} & \ours \scriptsize{(\cmark)} & LRP & IG & LeG & Chf & \ours \scriptsize{(\cmark)} & \\
\midrule

\multicolumn{11}{@{}l@{}}{\textit{Vision--language teacher: SigLIP2}~\cite{tschannen2025siglip}}\\
\addlinespace[2pt]
ViT-B/16      & 84.21 & 83.87 & 80.39 & 79.97 & 55.47 & 39.22 & 34.10 & 54.27 & \textbf{84.42} & \textcolor{teal}{+28.95} \\
ViT-L/16      & 87.15 & 87.08 & 83.76 & 83.89 & 47.39 & 37.46 & 36.62 & 52.57 & \textbf{83.48} & \textcolor{teal}{+36.09} \\
ViT-so/16     & 87.98 & 87.83 & 84.51 & 84.56 & 49.81 & 37.82 & 46.09 & 55.03 & \textbf{82.75} & \textcolor{teal}{+32.94} \\
ViT-so/14@384 & 88.65 & 88.48 & 85.08 & 85.13 & 49.54 & 37.71 & 45.18 & 56.03 & \textbf{83.07} & \textcolor{teal}{+33.53} \\

\midrule
\multicolumn{11}{@{}l@{}}{\textit{Self-supervised teacher: DINOv3}~\cite{simeoni2025dinov3}}\\
\addlinespace[2pt]
ViT-S/16      & 78.62 & 77.91 & 76.92 & 75.83 & 53.46 & 33.11 & 31.72 & 31.27 & \textbf{76.98} & \textcolor{teal}{+23.52} \\
ViT-B/16      & 84.36 & 84.14 & 82.28 & 81.56 & 65.11 & 37.61 & 36.61 & 35.94 & \textbf{87.01} & \textcolor{teal}{+21.90} \\
ViT-L/16      & 87.08 & 86.82 & 84.77 & 83.71 & 65.33 & 38.47 & 35.90 & 36.92 & \textbf{84.04} & \textcolor{teal}{+18.71} \\

\midrule
\multicolumn{11}{@{}l@{}}{\textit{Supervised teacher: ViT}~\cite{dosovitskiy2021an}}\\
\addlinespace[2pt]
ViT-B/16      & 81.19 & 81.30 & 80.70 & 80.76 & 57.34 & 44.43 & - & - & \textbf{82.48} & \textcolor{teal}{+25.14} \\

\bottomrule
\end{tabular}%
}
\caption{\textbf{Localization (Grid-PG) vs.\ recognition.} \ours improves Grid-PG substantially across backbones and teachers while maintaining competitive linear-probe and $k$-NN accuracy on ImageNet-1k. Grid-PG is reported for Teacher using AttnLRP (LRP) \cite{achtibat2024attnlrp}, Integrated Gradients (IG) \cite{sundararajan2017axiomatic}, LeGrad (LeG) \cite{bousselham2024legrad}, and CheferCAM (Chf) \cite{chefer2021transformer}, while \ours uses model inherent B-cos attributions. $\Delta_{\text{GridPG}}$ is \ours minus the teacher with AttnLRP (LRP). \cmark: denotes inherently interpretable models. Fields marked with - (LeGrad and CheferCAM on the supervised ViT) were not implemented for this architecture.}
\label{tab:gridpg_localization}
\end{table*}

\myparagraph{Localization (GridPG).}
\ours aligned models yield substantial improvements in GridPG localization across pre-training paradigms and backbones while maintaining competitive performance in both recognition and dense prediction tasks. Compared to conventional teacher models explained with post-hoc attribution methods (AttnLRP~\cite{achtibat2024attnlrp}, Integrated Gradients~\cite{sundararajan2017axiomatic}, LeGrad~\cite{bousselham2024legrad}, and CheferCAM~\cite{chefer2021transformer}), inherently interpretable \bcos explanations from \ours achieve consistently higher localization scores. The gains are substantial across SigLIP2, DINOv3, and supervised ViT backbones, with \textcolor{teal}{$\Delta_{\text{GridPG}}$} ranging from \textbf{+18.71} to \textbf{+36.09}, showing that aligned \bcos models provide more localized, class-specific attributions without sacrificing downstream accuracy.

\begin{figure}[h!]
  \centering
  \begin{subfigure}[b]{0.49\linewidth}
    \centering
    \includegraphics[width=\linewidth]{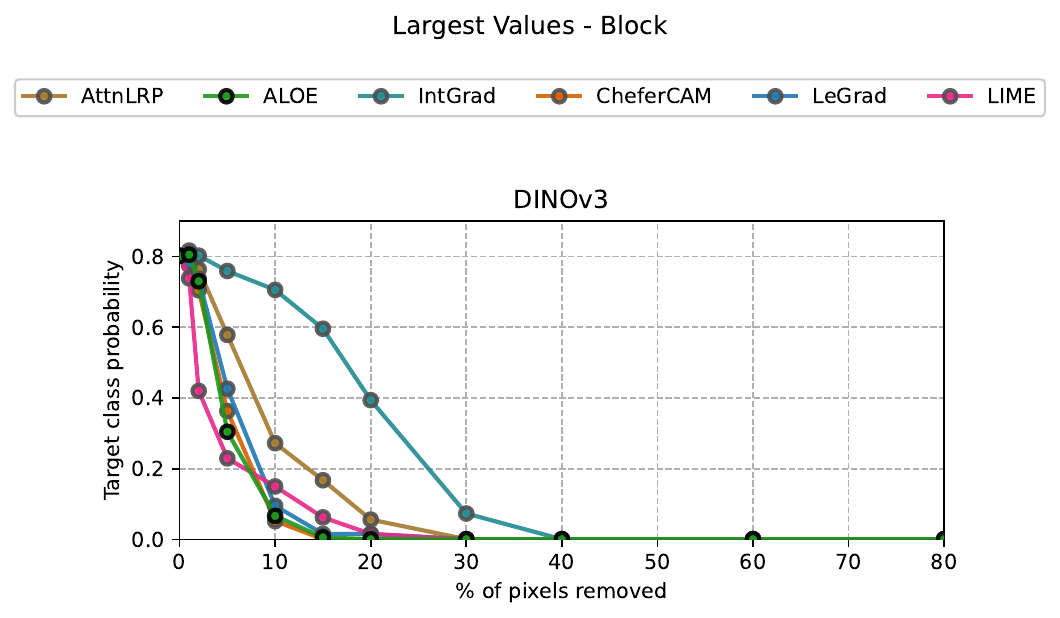}
    \label{fig:perturb_dinov3}
  \end{subfigure}
  \hfill
  \begin{subfigure}[b]{0.49\linewidth}
    \centering
    \includegraphics[width=\linewidth]{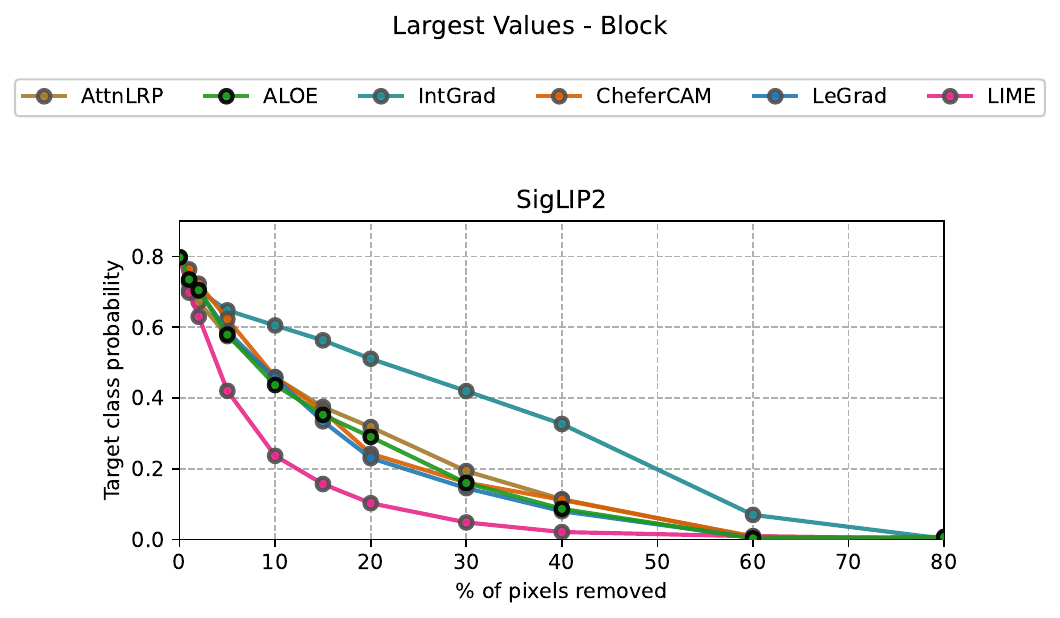}
    \label{fig:perturb_siglip2}
  \end{subfigure}
  
\caption{\textbf{Perturbation stability of explanations on ViT-B/16.} Target-class probability vs.\ percentage of top-attributed $16 \times 16$ pixel blocks removed (lower curves are better). Across both teachers—\textbf{DINOv3} (left) and \textbf{SigLIP2} (right)—removing the blocks ranked highest by \textbf{ALOE (ours)} model-inherent \bcos attributions $\mathbf{W}(\mathbf{x})\mathbf{x}$~\cite{boehle2024bcos} causes a rapid drop in model confidence, indicating faithful localization. Interestingly, LIME~\cite{ribeiro2016should} attributions cause the fastest initial drop for both teachers and perform best overall for SigLIP2, while ALOE remains competitive.
} 
  \label{fig:perturb_stability}
\end{figure}

\myparagraph{Faithfulness under pixel perturbation.}
In pixel-deletion tests on ViT-B/16 (\cref{fig:perturb_stability}), target-class probability for \ours (using model-inherent $\mathbf{W}(\mathbf{x})\mathbf{x}$) drops very fast when $16\times16$ pixel blocks with the highest attribution are removed, outperforming AttnLRP and Integrated Gradients for both DINOv3 and SigLIP2. Interestingly, LIME~\cite{ribeiro2016should} attributions cause the fastest initial drop for both teachers and perform best overall for SigLIP2, while ALOE remains competitive. The faster decay indicates more faithful attributions that better reflect the model’s decision computations. The usage of blocks instead of single pixels is necessitated by the inclusion of CheferCAM, LIME and LeGrad which work internally at token-granularity; thus providing less fine-grained attribution maps compared to B-cos, AttnLRP or IntGrad.

\myparagraph{Post-hoc methods increase localization score.} 
Applying the same post-hoc methods to the aligned \ours models (see Fig.~\ref{fig:gridpg_baseline_aloe_post_hoc}) substantially increases the GridPG score by up to \textbf{+27.0} p.p., which narrows the gap to the score achieved by the inherent explanation mechanism. This works especially well for CheferCAM, LeGrad, and IntGrad, while AttnLRP improves less because it already achieves relatively high scores on the standard teacher. This suggests that B-cos models generally increase object localization, independent of the attribution method (similar findings were also presented in~\cite{gairola2025probe}).
    
\begin{figure}[t]
  \centering
  \includegraphics[width=0.75\linewidth]{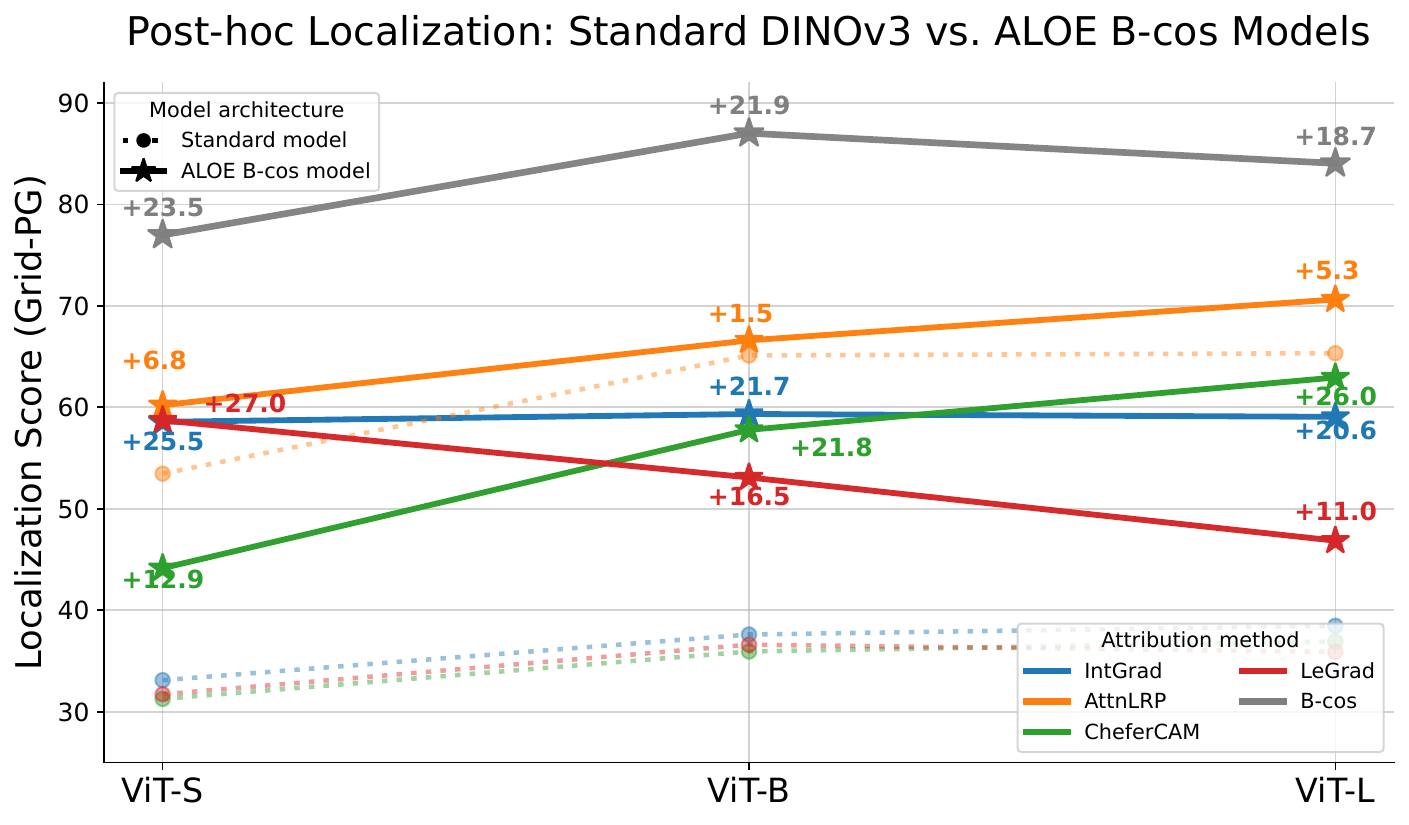}
  \caption{\textbf{GridPG on standard DINOv3 and \ours.} We compute GridPG scores on DINOv3 ViT-S/B/L teachers and their aligned \ours counterparts using the same post-hoc attribution methods. Applying post-hoc methods to \ours models increases localization strongly across most methods, narrowing the gap to inherent B-cos explanations.}
  \label{fig:gridpg_baseline_aloe_post_hoc}
\end{figure}

\myparagraph{Human user study.}
We also performed a human user study in which the subjective preferences of different attribution methods for DINOv3 ViT-B/16 were evaluated. 41 users on Amazon Mechanical Turk performed 50 trials, each consisting of a random image with 2 attribution maps from randomly drawn attribution methods (AttnLRP, IntGrad, Chefer, LeGrad, LIME, B-cos). An example for one trial is shown in \cref{fig:human_study_example}. The predicted class of the image was provided, and the user had to choose the explanation that best helped them understand the model's prediction. We fitted a Bradley–Terry (BT) model \cite{bradley1952rank} to the outcome of the study in order to derive the pairwise win-probabilities between attribution methods. BT is similar to logistic regression on the one-hot encoded study data, where the winning class receives the score $+1$ and the negative $-1$ (with $1$ as the target value). The derived BT weights indicate the overall performance relative to the other models. By computing the sigmoid of the difference $\beta_1-\beta_2$ between the weights of two attribution methods, we can derive the probability of method $1$ winning against method $2$. The probabilities, together with the BT-scores, are shown in \cref{tab:human_eval_styled}. ALOE is preferred over all post-hoc baselines ($>50\%$ win rate) while being the most compute-efficient method.
\\
\begin{figure}
    \centering
    \includegraphics[width=0.8\linewidth]{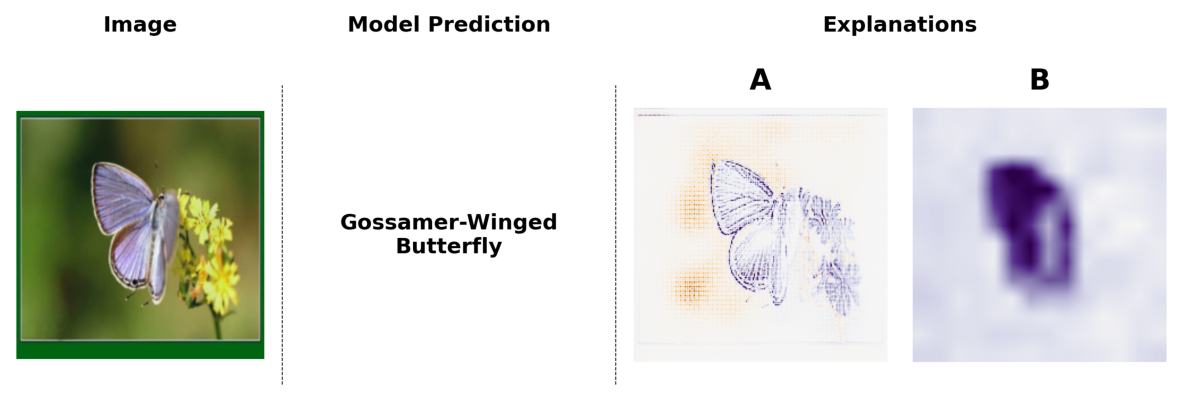}
    \caption{
    \textbf{Example for the User Study\prevreg.} Users were given two attribution maps from randomly drawn methods and had to decide which attribution map helps them best to understand the model's prediction. Blue shows positive attributions while yellow shows negative ones. \prevregnote
    } 
    \label{fig:human_study_example}
\end{figure}

\begin{table}[h]
\centering
\small
\caption{\textbf{Human Evaluation\prevreg.} \ours achieves the highest Bradley--Terry (BT) score and outperforms all baselines in pairwise comparisons (Win Rate $>50\%$). \prevregnote}
\label{tab:human_eval_styled}
\setlength{\tabcolsep}{10pt}
\begin{tabular}{@{}l cc@{}}
\toprule
\textbf{Method} & \textbf{BT Score} $\uparrow$ & \textbf{\ours Win Rate} $\uparrow$ \\
\midrule
\rowcolor{hirow}
\textbf{\ours\prevreg} & \textbf{\phantom{-}0.97} & \textbf{--} \\

IntGrad   & \phantom{-}0.41 & 63.5\% \\
AttnLRP   & \phantom{-}0.11 & 70.2\% \\
CheferCAM & -0.11           & 74.5\% \\
LeGrad    & -0.59           & 82.6\% \\
LIME      & -0.79           & 85.3\% \\
\bottomrule
\end{tabular}
\end{table}

\subsection{Ablations}
\label{sec:ablations}

We present the table on ablation for \emph{alignment objectives} as reported in ~\cref{sec:main_ablations} of the main paper.

\begin{table}[h!]
\centering
\caption{\textbf{Alignment objective ablation\prevreg.} IN1k top-1 (\%) using only the global alignment term ($\mathcal{L}_{\mathrm{global}}$, no layer-wise terms); the selected objective was then extended to intermediate depths (\cref{tab:where_loss_cols}). \prevregnote}
\label{tab:loss_ablation_in1k_lp}
\small
\setlength{\tabcolsep}{6pt}
\begin{tabular}{@{}l r r r r@{}}
\toprule
\textbf{Loss} & \textbf{MSE} & \textbf{Cosine} & \textbf{SigLIP} & \textbf{InfoNCE} \\
\midrule
DINOv3     & 83.9 & \textbf{84.0} & 83.7 & 83.5 \\
SigLIP2    & 75.5 & 75.8 & \textbf{76.0} & 75.7 \\
Google ViT & 81.0 & \textbf{81.1} & \textbf{81.1} & 81.0 \\
\bottomrule
\end{tabular}
\end{table}

\myparagraph{Alignment objectives.}
Under identical settings with only the global alignment term (Tab.~\ref{tab:loss_ablation_in1k_lp}), \textit{cosine} and \textit{SigLIP} are most consistent across models, \textit{MSE} and \textit{InfoNCE} showed inconsistencies. Given the ease of application and simplicity we adopt \textbf{cosine} by default.

\clearpage
\section{Additional Qualitative Results}
\label{supp:sec:qualitative}

In this section, we complement the quantitative results with qualitative evidence across three settings: (i) \emph{zero-shot, model-inherent} explanations (\Cref{fig:zs_vlm_explanations}), (ii) side-by-side comparisons against popular post-hoc attribution methods (\Cref{fig:qual_attrib_1,fig:qual_attrib_2,fig:qual_attrib_bad_examples}), (iii) dense predictions from a linear depth probe (\Cref{fig:depth_map}), and multimodal large language model explanations (\cref{fig:vlm_teaser_supplement_1,fig:vlm_teaser_supplement_2,fig:vlm_teaser_supplement_3}).

\myparagraph{Zero-shot, model-inherent VLM explanations.}
In \Cref{fig:zs_vlm_explanations}, we visualize zero-shot predictions by swapping the SigLIP2 image encoder with its ALOE-aligned \bcos counterpart while keeping the original text encoder and prompts. The resulting \emph{inherent} explanations localize the class-relevant regions (e.g., discriminative parts, textures) without any additional tuning. Notably, maps remain well-aligned and class-specific, consistent with our zero-shot accuracy in \cref{tab:zeroshot_siglip2}.

\begin{figure}[t]
  \centering
  \includegraphics[width=\linewidth]{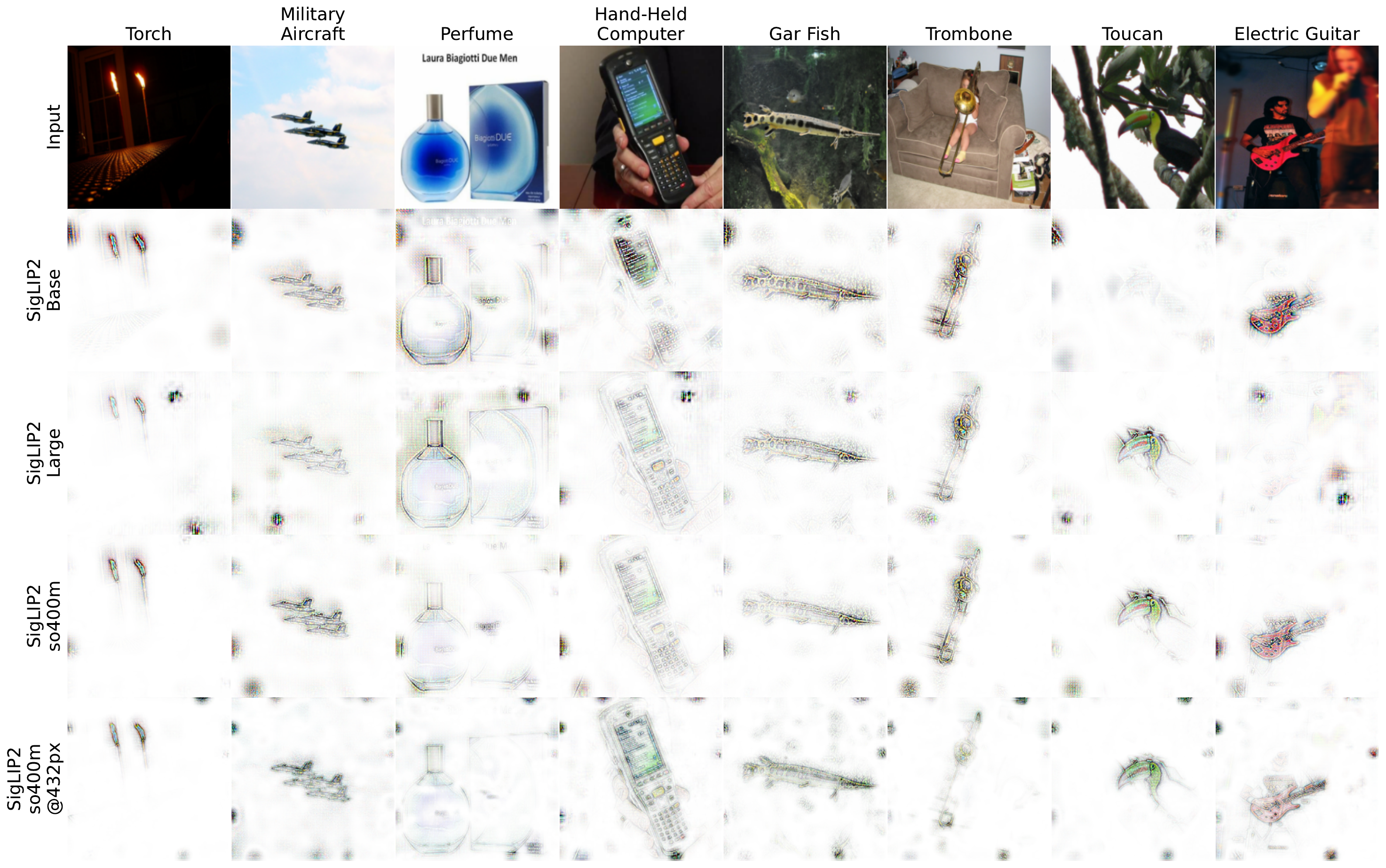}
  \caption{\textbf{Zero-shot, model-inherent VLM explanations.} RGB explanations of \textbf{ALOE (ours)} using \(\mathbf{W}(\mathbf{x})\mathbf{x}\) attributions, given the fixed text prompt \emph{``this is a photo of \{class name\}''}. Our explanations are sharply localized on class-relevant regions.}
  \label{fig:zs_vlm_explanations}
\end{figure}

\myparagraph{Comparisons with popular attribution methods.}
\Cref{fig:qual_attrib_1,fig:qual_attrib_2,fig:qual_attrib_bad_examples} contrast \textbf{ALOE (ours)}—which uses model-inherent \bcos attributions $\mathbf{W}(\mathbf{x})\mathbf{x}$—with AttnLRP, Integrated Gradients, LeGrad, CheferCAM, and LIME. Across diverse categories, \ours produces more object-centric explanations with sharper boundaries and less background noise. The six-channel encoding preserves color semantics, yielding explanations that align with class-specific parts and textures. These trends also mirror our quantitative gains in GridPG and pixel-perturbation stability (\cref{tab:gridpg_localization,fig:perturb_stability}). 
\noindent \emph{All examples for this use DINOv3-based backbones (ViT-B/16) with linear probes trained on ImageNet-1k.}

\begin{figure}[t]
  \centering
  \includegraphics[width=\linewidth]{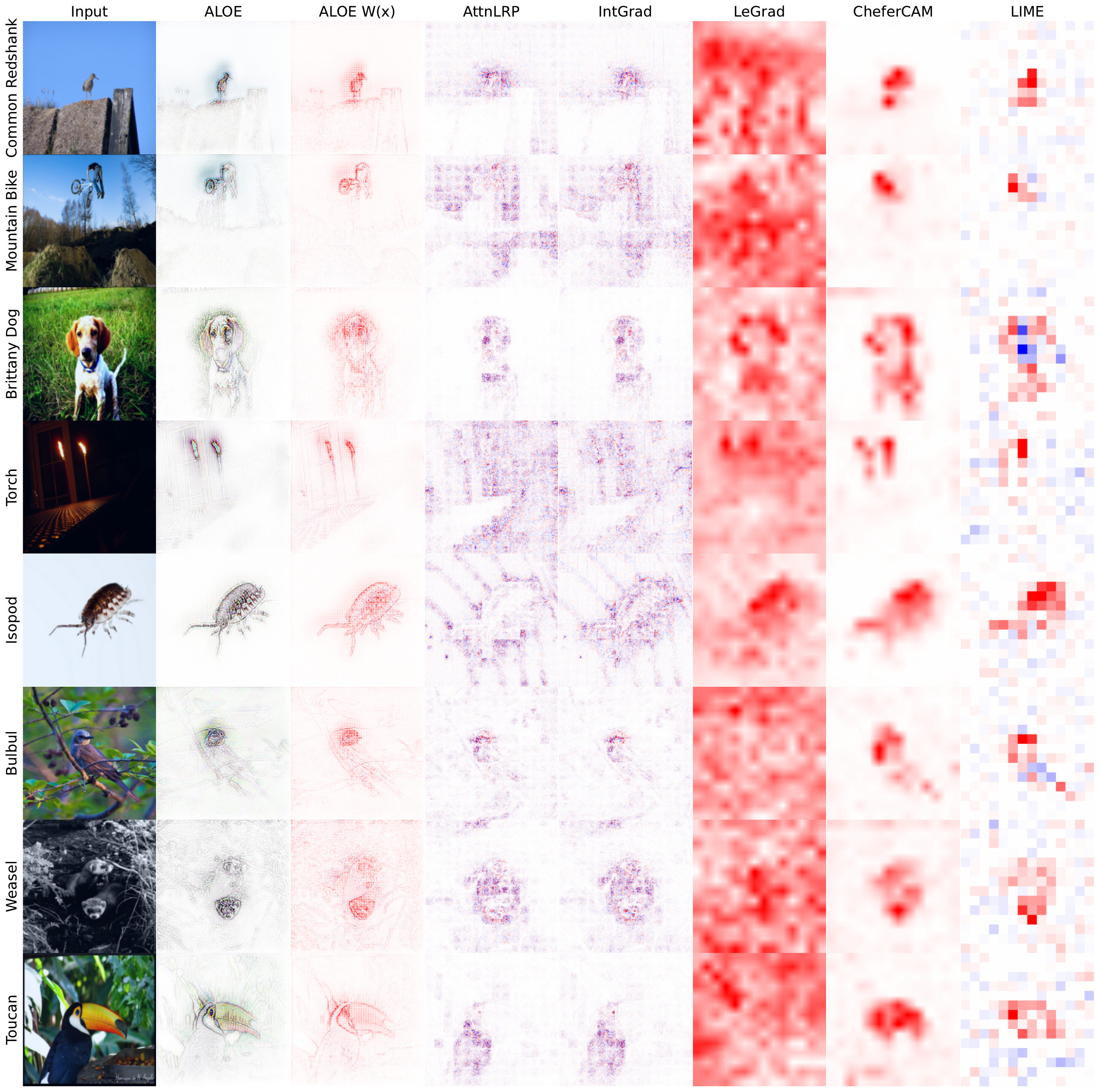}
  \caption{\textbf{Qualitative attribution comparisons.} Example 1:
  Visualizations for \textbf{ALOE (ours)}—using model-inherent B-cos attributions $\mathbf{W}(\mathbf{x})\mathbf{x}$—versus popular post-hoc methods (AttnLRP, Integrated Gradients (IntGrad), LeGrad, CheferCAM, and LIME).
  \ours produces sharper, better-localized, and color-faithful highlight maps with less background noise, focusing on class-relevant object regions consistently across examples.
  }
  \label{fig:qual_attrib_1}
\end{figure}

\begin{figure}[t]
  \centering
  \includegraphics[width=\linewidth]{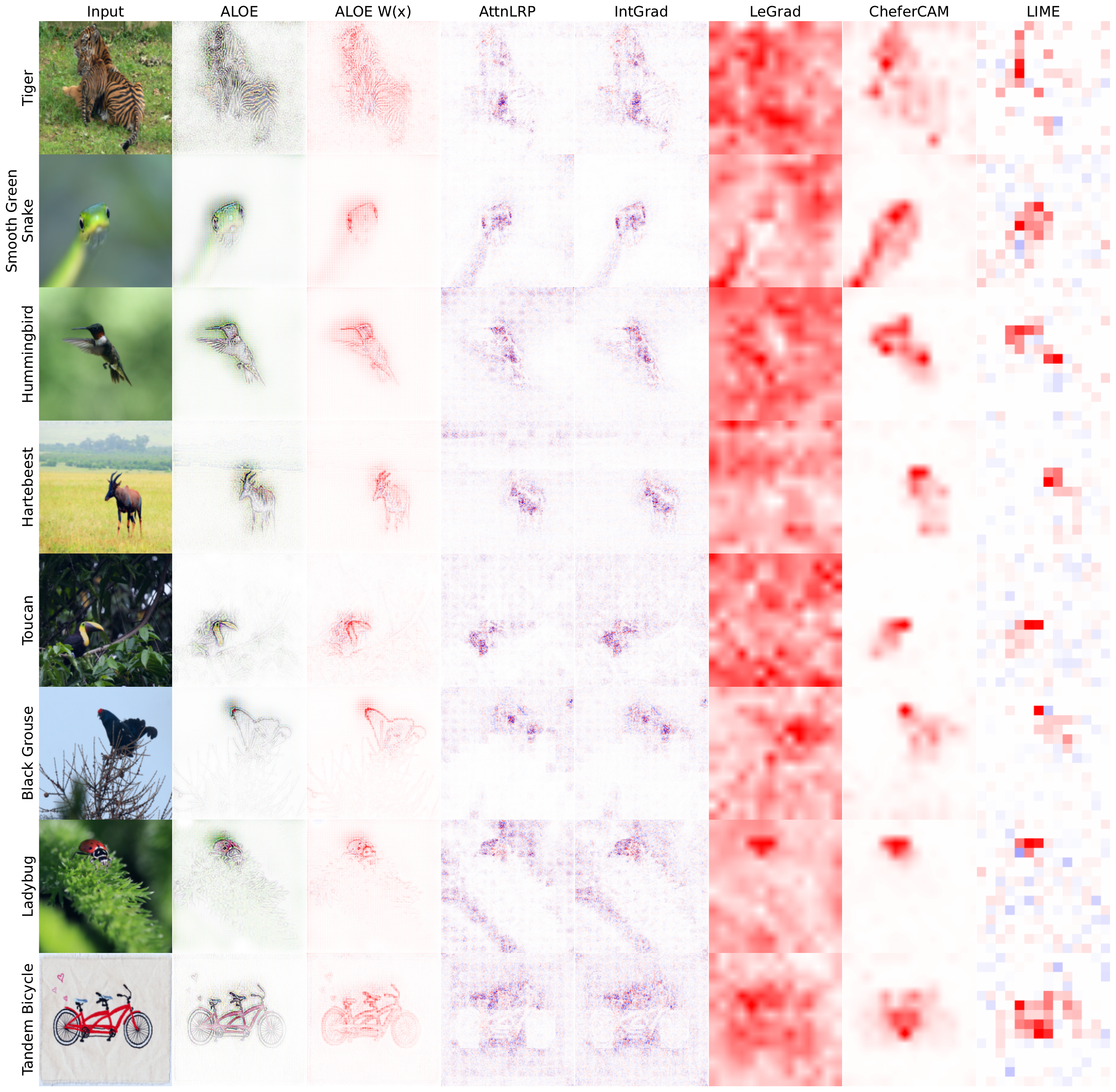}
  \caption{\textbf{Qualitative attribution comparisons.}
  Similar to Figure~\ref{fig:qual_attrib_1}.
  }
  \label{fig:qual_attrib_2}
\end{figure}

\begin{figure}[t]
  \centering
  \includegraphics[width=\linewidth]{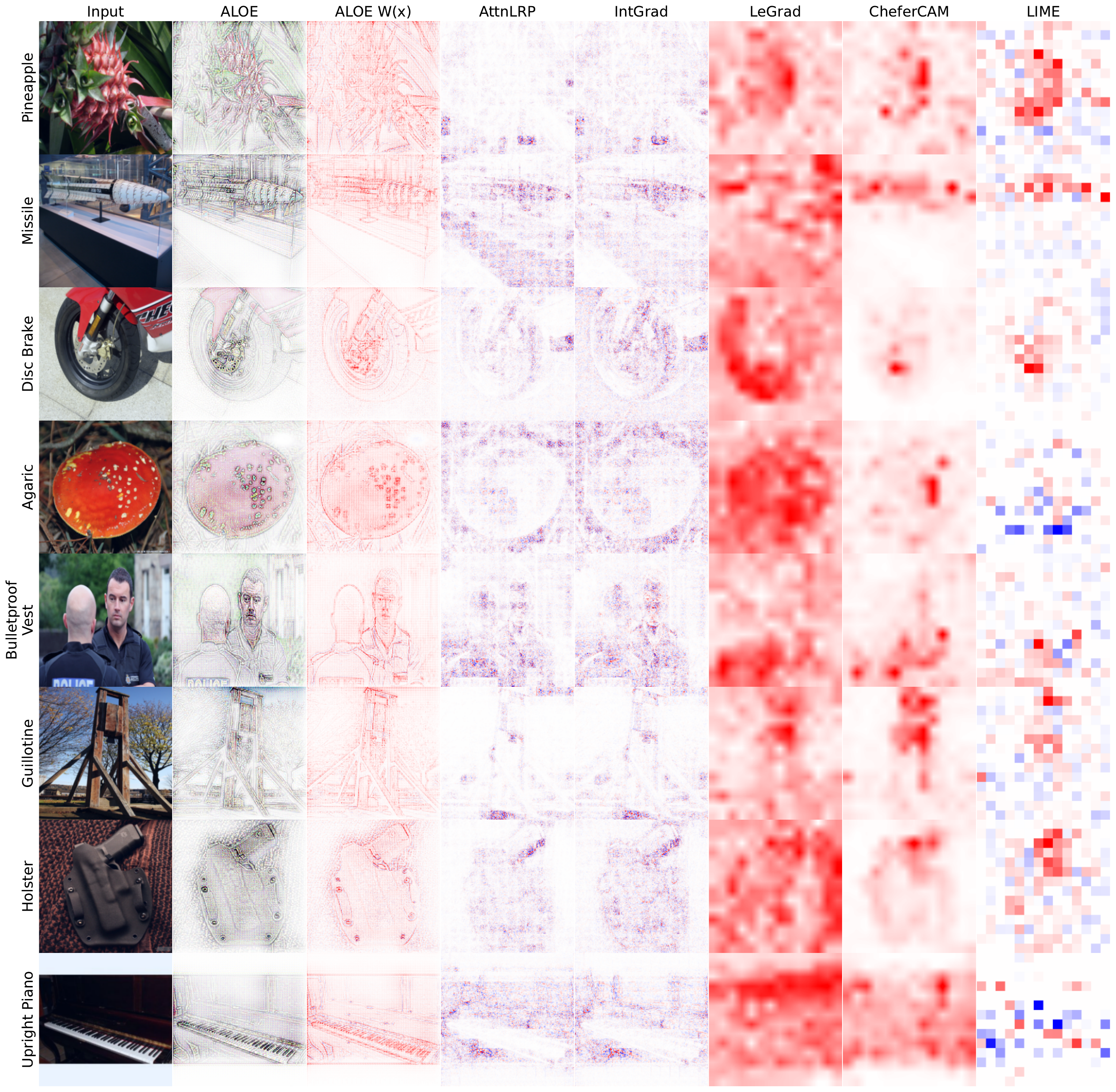}
  \caption{\textbf{Qualitative attribution comparisons--not well-localized examples.} Visualizations for \textbf{ALOE (ours)}—using model-inherent B-cos attributions $\mathbf{W}(\mathbf{x})\mathbf{x}$—versus popular post-hoc methods (AttnLRP, Integrated Gradients (IntGrad), LeGrad, CheferCAM, and LIME).
    We highlight failure cases where explanations lack localization, suggesting a reliance on background context. Given that B-cos attributions are mathematically faithful to the model’s linear transformation, these diffuse heatmaps could also expose model-level shortcuts rather than ``explanation failure."}
  \label{fig:qual_attrib_bad_examples}
\end{figure}

\clearpage
\begin{figure}[t]
  \centering
  \includegraphics[width=\linewidth]{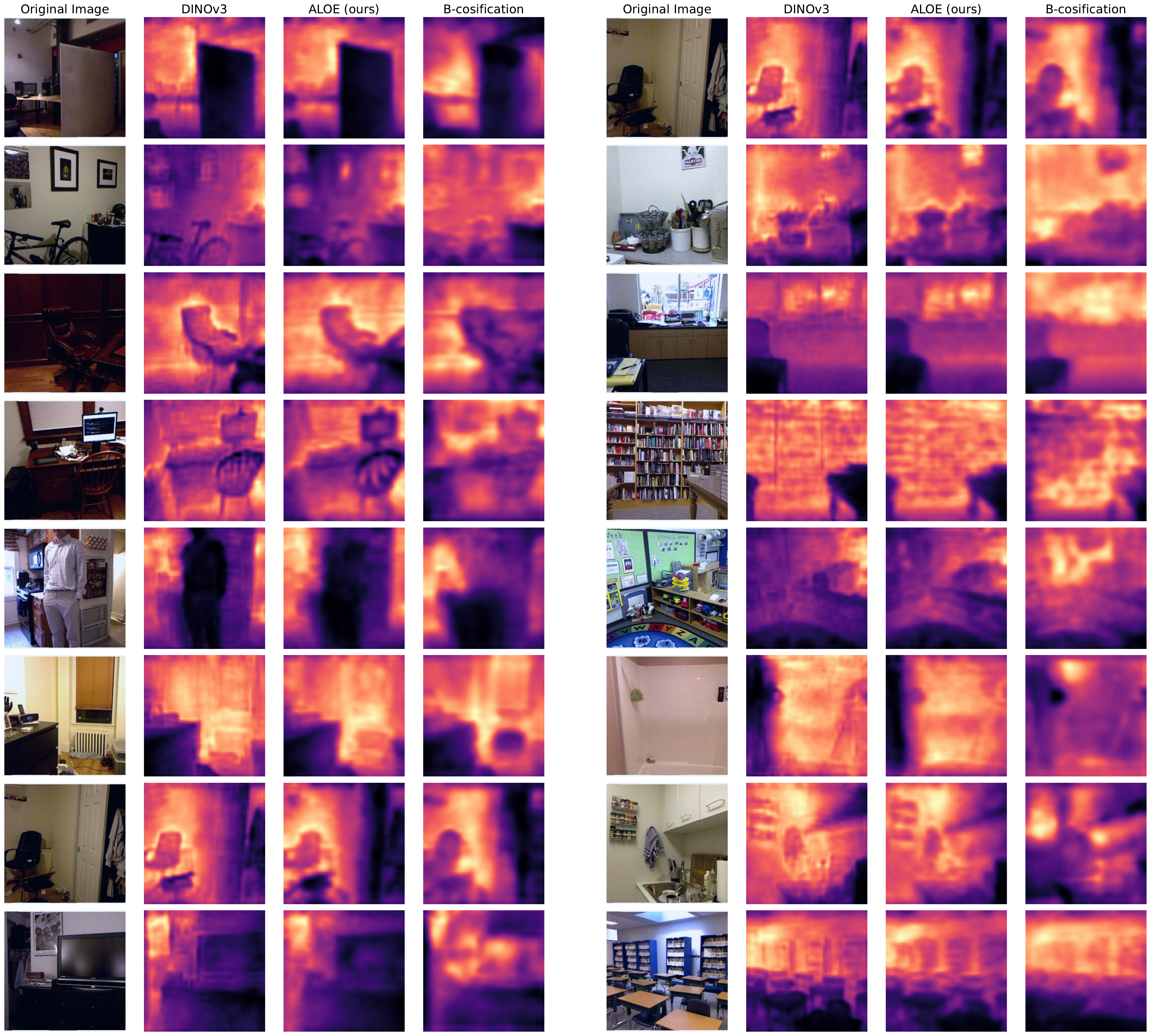}
  \caption{\textbf{Depth Estimation.} Visualization of predicted depth maps for \textbf{ALOE (ours)}, DINOv3 and B-cosification on the NYUv2 \cite{silberman2012indoor} depth-estimation dataset using the Probe3D \cite{el2024probing} protocol. \ours-aligned features yield depth maps with coherent geometry that visually are very similar to the DINOv3 teacher model. The depth maps from the vanilla B-cosification model seem to be more noisy and blurred.}
  \label{fig:depth_map}
\end{figure}

\myparagraph{Depth maps (dense linear probing).}
\Cref{fig:depth_map} shows monocular depth outputs from a shallow linear head trained on frozen features (ViT-B/16). ALOE-aligned features yield depth maps with coherent geometry that are visually very similar to the DINOv3 teacher. These visuals complement the relative/absolute depth metrics in \cref{tab:dense_depth_vitb16}, underscoring that the aligned \bcos backbone provides useful \emph{dense} representations—not only global classification signals—while retaining inherent interpretability. The depth maps from the vanilla B-cosification model seem to be noisier and more blurred.

\myparagraph{Multimodal large language model explanations.}
In addition to the explanations in \cref{fig:llava_grounding}, we provide further examples—including failure cases—in \cref{fig:vlm_teaser_supplement_1,fig:vlm_teaser_supplement_2,fig:vlm_teaser_supplement_3}. Using AttnLRP \cite{achtibat2024attnlrp}, we propagate relevance through the LLaVA-MORE \cite{cocchi2025llava} Gemma-2-9B \cite{gemmateam2024gemma2improvingopen} backbone, utilizing B-cos inherent explanations from the \ours SigLIP2 vision encoder. Notably, \cref{fig:vlm_teaser_supplement_2} and \cref{fig:vlm_teaser_supplement_3} demonstrate instances where explanations fail to match predictions. We suspect that this discrepancy likely stems from three factors: the model relying on spurious associations, reliance on previous tokens restricting the vision-to-language information flow, or AttnLRP fundamentally generating unfaithful explanations. As we discussed earlier, since the current setup still relies on AttnLRP to propagate relevance through the language model, we defer a fuller treatment of end-to-end inherently interpretable MLLMs to future work.

\begin{figure}[h!]
    \centering
    \includegraphics[width=0.85\linewidth]{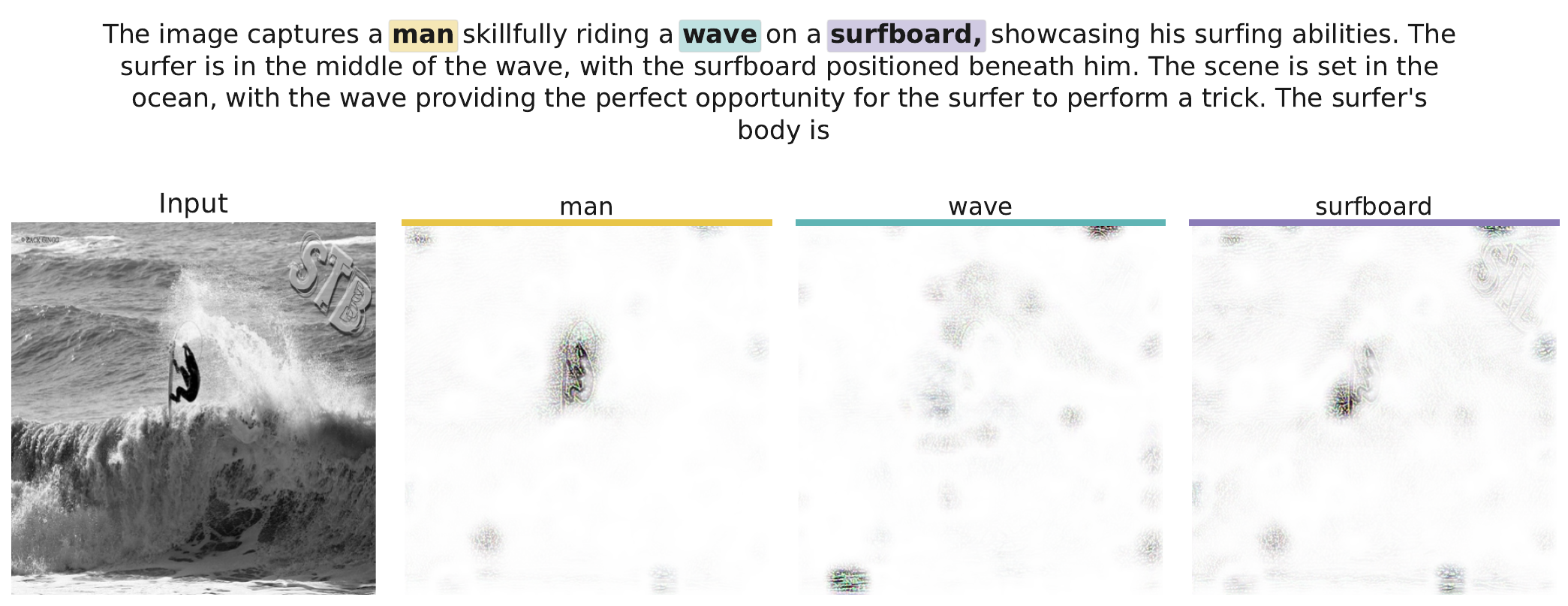}
    \caption{
    \textbf{Token-level Visual Grounding via AttnLRP and ALOE\prevreg.} We generate visual explanations by leveraging the inherent B-cos interpretability of the \ours\prevreg SigLIP2 vision backbone, using AttnLRP to propagate relevance from LLaVA-MORE's Gemma-2-9B language model. \prevregnote
    }
    \label{fig:vlm_teaser_supplement_1}
\end{figure}

\begin{figure}
    \centering
    \includegraphics[width=0.85\linewidth]{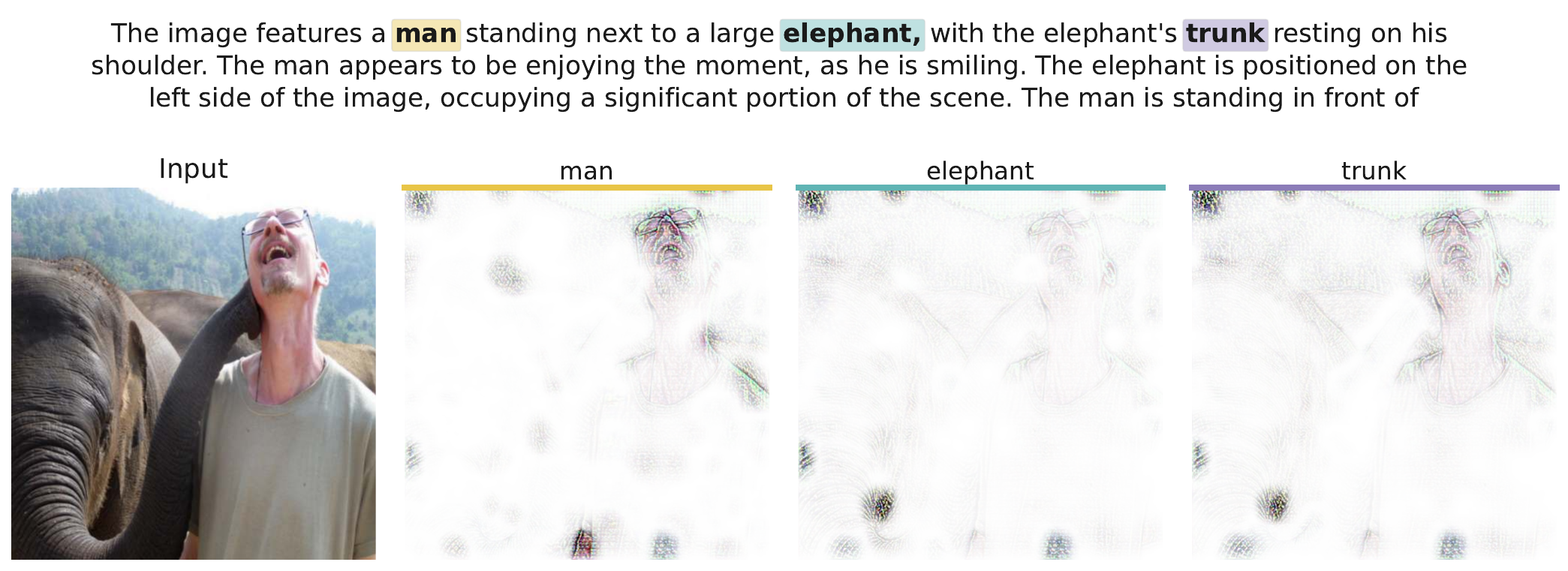}
    \caption{
    \textbf{Token-level Visual Grounding via AttnLRP and ALOE\prevreg.} Similar to \cref{fig:vlm_teaser_supplement_1}. \prevregnote
}
    \label{fig:vlm_teaser_supplement_2}
\end{figure}

\begin{figure}
    \centering
    \includegraphics[width=0.85\linewidth]{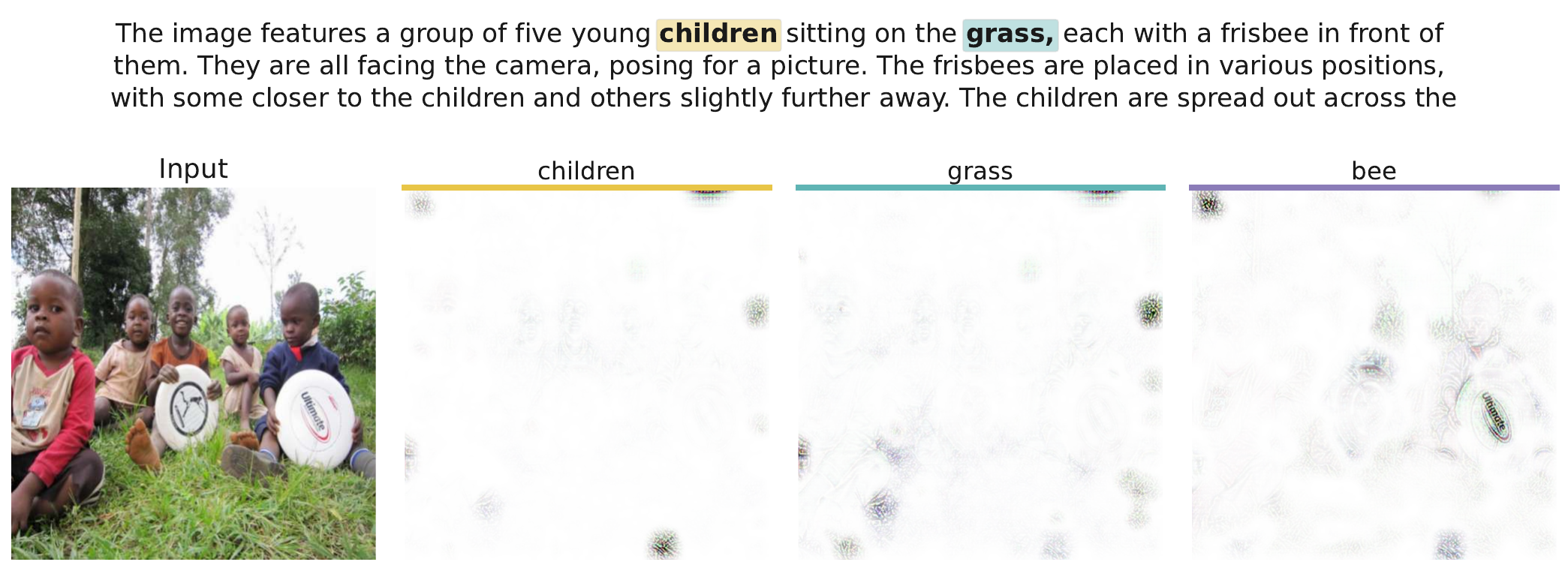}
    \caption{
    \textbf{Token-level Visual Grounding via AttnLRP and ALOE\prevreg.} Similar to \cref{fig:vlm_teaser_supplement_1,fig:vlm_teaser_supplement_2}. \prevregnote
}
    \label{fig:vlm_teaser_supplement_3}
\end{figure}

\clearpage
}

\end{document}